\pdfoutput=1

\documentclass[11pt]{article}
\usepackage[dvipsnames]{xcolor}

\usepackage[preprint]{acl}

\usepackage{times}
\usepackage{latexsym}

\usepackage[T1]{fontenc}

\usepackage[utf8]{inputenc}

\usepackage{microtype}

\usepackage{inconsolata}

\usepackage{graphicx}
\usepackage{tabularx}
\usepackage{multirow}
\usepackage{amsfonts}
\usepackage{amsmath}
\usepackage{xspace}
\usepackage{lipsum}
\usepackage{tcolorbox}
\usepackage{adjustbox}
\usepackage{booktabs}
\usepackage{makecell}
\usepackage{url}
\usepackage{enumitem}
\setlist[enumerate]{topsep=0pt, partopsep=0pt, parsep=0pt, itemsep=0pt, leftmargin=*}
\setlist[itemize]{topsep=0pt, partopsep=0pt, parsep=0pt, itemsep=0pt, leftmargin=*}
\usepackage{pgfplots}
\pgfplotsset{compat=1.18}
\usepackage{listings}
\usepackage{subcaption}
\usepackage{adjustbox}
\usepackage{pgf-pie}
\usepackage{subcaption}

\lstdefinestyle{mypython}{
  language=Python,
  basicstyle=\ttfamily\footnotesize, 
  keywordstyle=\bfseries\color{blue},
  stringstyle=\color{orange},
  commentstyle=\color{gray},
  numbers=left,
  numberstyle=\tiny\color{gray},
  stepnumber=1,
  numbersep=5pt,
  frame=single,
  breaklines=true,           
  breakatwhitespace=true,    
  columns=fullflexible,      
  keepspaces=true,
  showstringspaces=false,
  tabsize=2,
  captionpos=b,
  xleftmargin=0pt,           
  xrightmargin=0pt,
  linewidth=\columnwidth     
}

\newcommand{\lap}{$\textit{LLM-as-planner}$\xspace}
\newcommand{\laf}{$\textit{LLM-as-formalizer}$\xspace}

\newcommand{\penn}{%
  \hspace{1pt}
  \begingroup\normalfont
  \includegraphics[height=1.3\fontcharht\font`\B]{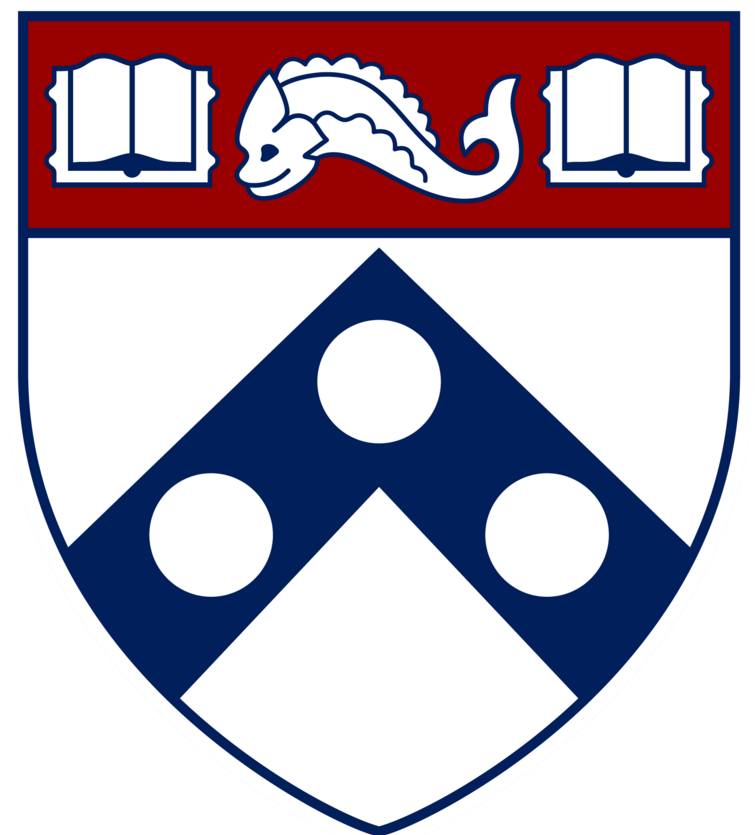}%
  \endgroup
  \hspace{1pt}
}
\newcommand{\drexel}{%
  \hspace{1pt}
  \begingroup\normalfont
  \includegraphics[height=1.3\fontcharht\font`\B]{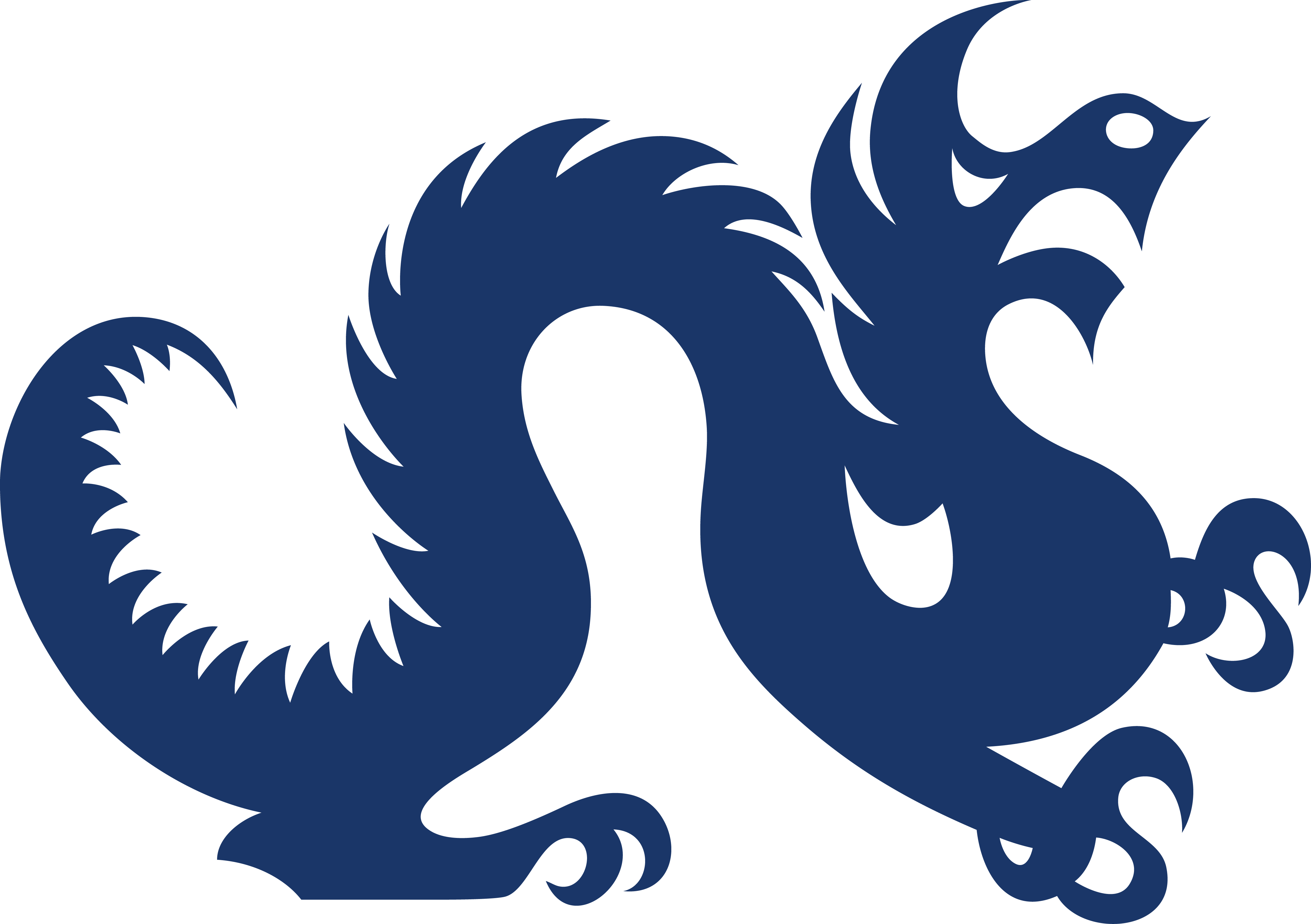}%
  \endgroup
  \hspace{1pt}
}
\newcommand{\jhu}{%
  \hspace{1pt}
  \begingroup\normalfont
  \includegraphics[height=1.3\fontcharht\font`\B]{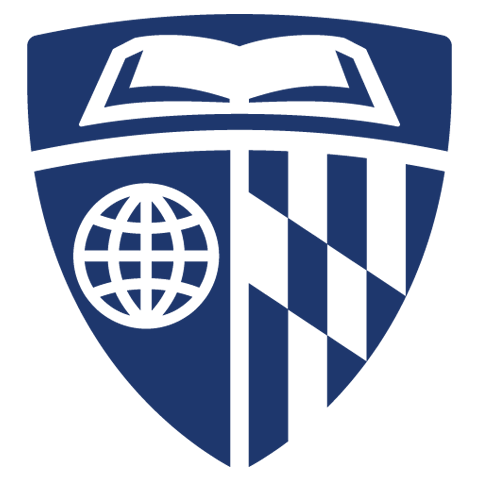}%
  \endgroup
  \hspace{1pt}
}
\newcommand{\nw}{%
  \hspace{1pt}
  \begingroup\normalfont
  \includegraphics[height=1.3\fontcharht\font`\B]{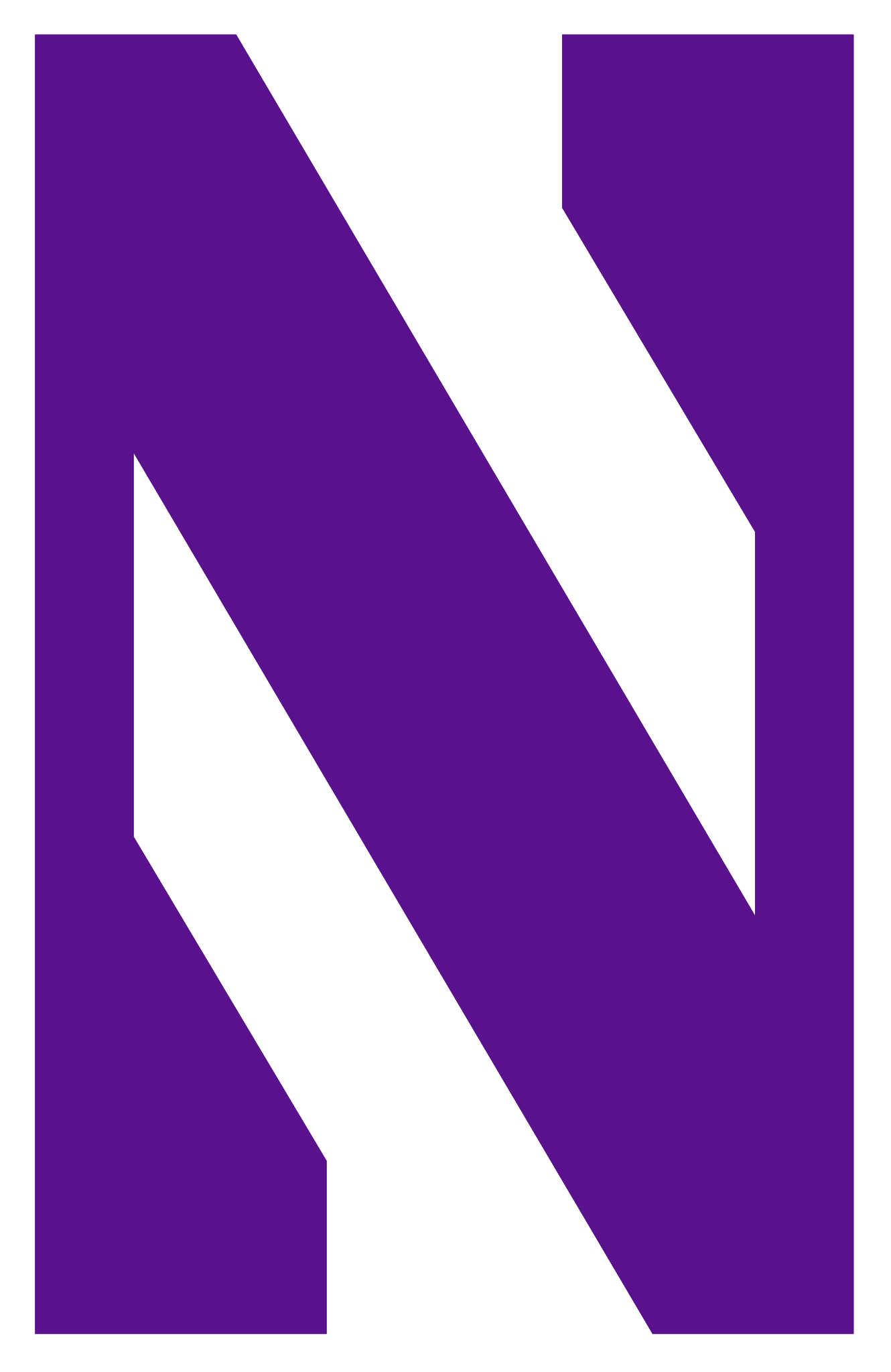}%
  \endgroup
  \hspace{1pt}
}

\title{Unifying Inference-Time Planning Language Generation}

\author{
  Prabhu Prakash Kagitha\drexel \quad
  Bo Sun\penn \quad
  Ishan Desai\drexel \quad
  Andrew Zhu\penn \\
  \textbf{Cassie Huang}\drexel \quad
  \textbf{Manling Li}\nw \quad
  \textbf{Ziyang Li}\jhu \quad
  \textbf{Li Zhang}\drexel \\
  \drexel Drexel University \quad
  \penn University of Pennsylvania \\
  \nw Northwestern University \quad
  \jhu Johns Hopkins University \\
  \texttt{\{prabhuprakash.k,harry.zhang\}@drexel.edu}
}

\begin{document}

\maketitle

\begin{abstract}
A line of work in planning uses LLM not to generate a plan, but to generate a formal representation in some planning language, which can be input into a symbolic solver to deterministically find a plan. While showing improved trust and promising performance, dozens of recent publications have proposed scattered methods on a variety of benchmarks under different experimental settings. We attempt to unify the inference-time LLM-as-formalizer methodology for classical planning by proposing a unifying framework based on intermediate representations. We thus systematically evaluate more than a dozen pipelines that subsume most existing work, while proposing novel ones that involve syntactically similar but high resource intermediate languages (such as a Python wrapper of PDDL). We provide recipes for planning language generation pipelines, draw a series of conclusions showing the efficacy of their various components, and evidence their robustness against problem complexity.\footnote{Our code and data are at \url{https://github.com/prakashkagitha/llm-pddl-gen}.}
\end{abstract}

\begin{figure*}
    \centering
    \includegraphics[width=\textwidth]{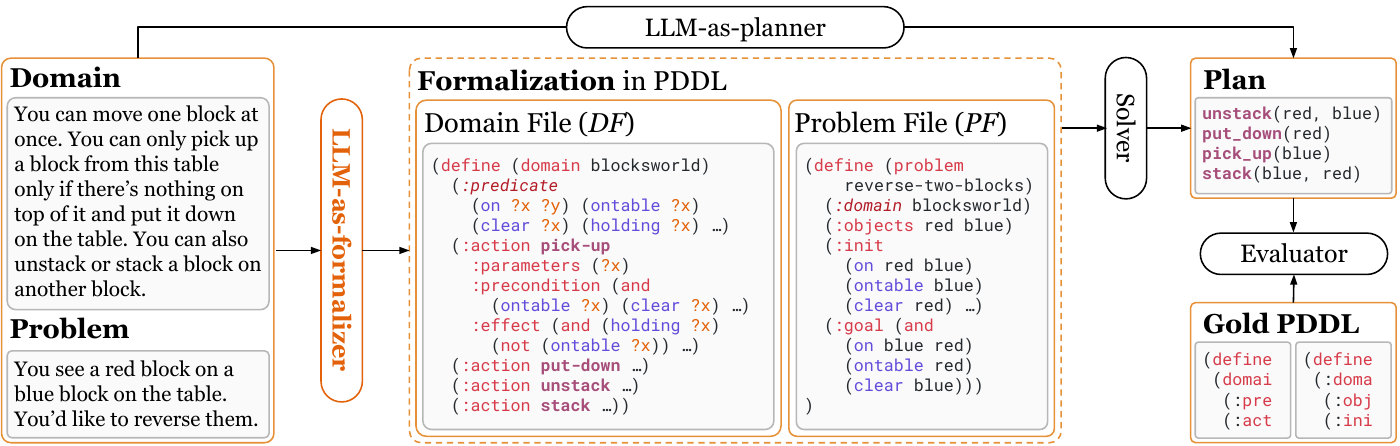}
    \caption{An illustration of using LLM as a planner or a formalizer in classical planning. While \lap generates a plan directly, \laf formalizes a PDDL domain file and problem which evoke a solver to find a plan. The plan is evaluated against ground-truth PDDL that simulates the environment.}
    \label{fig:planner_formalizer}
\end{figure*}

\section{Introduction}

Large language models (LLMs) have been extensively applied to classical planning, an important step for complex problem solving in artificial intelligence. 
Given a textual description of the domain and problem of an environment, LLMs often directly generate a sequence of actions in an end-to-end manner \citep{wei-etal-2025-plangenllms}. Despite much better domain adaptation ability than symbolic methods, the \lap methodology is not verifiable and interpretable by nature, in addition to underperforming complex problems \citep{valmeekam2023planbench}.  
Alternatively, a recent line of work re-imagines the role of LLMs not to generate the plan, but to generate a formal program in languages such as the Planning Domain Definition Language (PDDL). Such a program is input into a formal solver to deterministically search for a plan \citep{tantakoun-etal-2025-llms}. The \laf methodology (Figure~\ref{fig:planner_formalizer}) has been widely advocated in literature due to its reportedly better performance and formal guarantees compared to \lap \citep{liu2023llm+}.
However, the success of \laf in previous work has primarily been tied to closed-source LLMs that are likely huge in terms of parameters and esoteric in terms of engineering. In contrast, preliminary experiments on open-source, smaller LLMs have been shown to exhibit much lower performance \citep{huang-zhang-2025-limit}, severely hindering the accessibility and democratization of LLM-assisted planning. Moreover, most if not all past work has employed unsystematic designs of pipelines of piecemeal techniques, hence the direction of improvement remains unclear.

In this work, we attempt to unify the \laf methodology for classical planning. We first propose a theoretical framework based on the number and interplay of intermediate representations (IR). This framework subsumes most if not all prior work on planning language generation while opening new paths for experimentation. Concretely, we propose the use of a high-resource but syntactically similar IR, namely a Python wrapper of PDDL, to achieve competitive performance. We systematically instantiate and evaluate pipelines that subsume most prior work and validate the challenge for open-source, mid-sized LLMs no more than 32B parameters. On 4 standard benchmarks \cite{ipc}, we show that IR and revision by solver feedback are the keys ingredients of a successful \laf pipeline, which consistently outperforms \laf especially when the problem complexity increases. Even so, other intuitive methods such as constrained decoding by grammar or a natural language IR resembling chain-of-thought do not improve performance over end-to-end PDDL generation. 
We pose this work as a building block to greatly accelerate progress in \laf and planning.




\section{Task Formulation}
\label{sec:formulation}

We will first formally define the classical planning task in line with established literature of STRIPS \citep{10.5555/1622876.1622939} and recent conventions in the natural language processing community \citep{huang-zhang-2025-limit}, before introducing our theoretic framework for \laf. 

Formally, the input $I$ of a classical planning task consists of a triplet $(D_d, D_p, \mathcal{DF_G'})$, where:
\begin{enumerate}
    \item $D_d$ is a textual description of the domain, specifying the types (the type system for entities $e$), predicates (in relation to the types), and actions $A$ including pre-conditions and effects (parameterized by conjunctive logical formulas over the the predicates).
    \item $D_p$ is a textual description of the problem, specifying the objects, initial states and goal states in terms of predicates;
    \item $\mathcal{DF_G'}$ is the header of the ground-truth PDDL domain file $\mathcal{DF_G}$ that only provides the names and parameters of the actions. This is required so that the eventual plan is grounded to an existing ontology and can be evaluated fairly.
\end{enumerate}

The final objective is a plan $L = [a_1(\bar{e}_1), \ldots, a_m(\bar{e}_m)]$, where each 
$a_i \in A$ is an action provided by $\mathcal{DF_G'}$, and 
each $\bar{e}_i$ is a tuple of grounded entities corresponding to the parameters of $a_i$. 
The plan $L$, when executed in the environment expressed by a ground-truth $\mathcal{DF_G}$ and $\mathcal{PF_G}$, must (i) be executable, fulfilling the pre-conditions of each action, and (ii) constitutes a successful transition from the initial states and goal states.

The domain file $\mathcal{DF}$ is defined by three main components:
types ($T$), 
predicates ($R$), and 
action semantics ($a_m, a_p,a_e$):
\begin{enumerate}
    \item Types: $T = {t_1, \ldots, t_n}$ is the set of entity types present in the environment.
    \item Predicates: $R = {r_1, \ldots, r_n}$ is the set of relational predicates, each in the form of $r(\bar{t})$, where $\bar{t}$ is a tuple of types that participate in relation $r$.
    \item Action semantics: given an action $a \in A$, $a_m$, $a_p$ and $a_e$ are its parameters, pre-conditions and effects. The parameters $a_m$ are a set of types $\bar{t}$ that participate in this action. The pre-conditions $a_p$ are a set of predicates $\bar{r}$ that must hold true for this action to be executed. The effects $a_e$ are a set of new predicates $\bar{r}'$ of which the new state takes the value after the execution of this action.
\end{enumerate}

The problem file $\mathcal{PF}$ is defined by three main components: 
objects ($E$), 
the initial state ($s_0$), and 
the goal state ($\psi_{\texttt{goal}}$):
\begin{enumerate}
    \item Objects: $E = {e_1, \ldots, e_n}$ is the set of named entities present in the environment. Each object is an instance of an entity type defined in $\mathcal{DF_G'}$.
    \item Initial State: $s_0 \in \mathcal{S}$ is a set of relational facts of the form $r(\bar{e})$, where $r \in R$ is a relational predicate defined in $\mathcal{D}$; $\bar{e}$ is a tuple of entities from $E$ that participate in relation $r$. 
    \item Goal State: $\psi_{\texttt{goal}}: \mathcal{S} \rightarrow \mathbb{B}$ is a boolean formula over relational facts, where $S$ is the state space and $\mathbb{B} = \{\text{True}, \text{False}\}$. When $\psi_{\texttt{goal}}(s^*) = \text{True}$, state $s^*$ achieves the problem goal.
\end{enumerate}

Here, the state space $\mathcal{S}$ comprises all possible configurations of relational facts over the object set $E$ with the predicates $R$. 
Each state $s \in \mathcal{S}$ is a set of instantiated facts $r(\bar{e})$, describing which relations hold among the objects.
Successfully executing a plan $L$ from the initial state $s_0$ yields a sequence of intermediate states $s_1, s_2, \ldots, s^*$ such that the goal formula $\psi_{\texttt{goal}}(s^*)$ evaluates to $\text{True}$.

While an end-to-end \lap pipeline maps the input triplet $I$ directly to a plan $L$, in this work we focus on \laf pipelines which instead generate the domain file $\mathcal{DF_P}$ and $\mathcal{PF_P}$ before inputting them into a PDDL solver to search for a plan. 

\section{Framework and Related Work}
\label{sec:framework}

We propose a theoretical framework for LLM-based classical planning revolving around levels of intermediate representations (IR), where the numbering of levels corresponds to the number of IRs involved.

\begin{itemize}
    \item Level 0: $I \rightarrow L$
    \item Level 1: $I \rightarrow$ PDDL $\leadsto L$
    \item Level 2: $I \rightarrow$ IR $\rightarrow$ PDDL $\leadsto L$
    \item Level 3: $I \rightarrow$ IR $\rightarrow$ IR $\rightarrow$ PDDL $\leadsto L$
    \item Level 4: $\dots$
\end{itemize}

Level 0 is essentially \lap, mapping the input $I$ to the plan $L$. While it is possible to involve IR in this level as a chain-of-thought, we do not discuss it as our focus is on \laf represented by the other levels, where PDDL is always the final IR before the plan. We can thus increase the level of complexity of a pipeline with more IRs. To instantiate a pipeline with the formulation above, we will first discuss the choice of IRs before the choice of modules.


\paragraph{Choice of IR}
IRs used in past work include:
\begin{enumerate}
    \item A \textbf{natural language}, where the LLM converts the input description into a free-form representation that facilitate later PDDL generation. This representation can range from loose sentences \citep{zhang-etal-2024-proc2pddl,hu-etal-2025-text2world} to a semi-formal data structure \citep{wong2024learning,hao2025planningrigorgeneralpurposezeroshot}, but in any case cannot be executed by an external solver.   
    Such a conversion can involve information extraction, chain-of-thought, or summarization. 
    \item Another \textbf{PDDL}, where the module that takes one PDDL as input and outputs another is a revision module discussed in details below.
\end{enumerate}

We will show in later discussions that neither of these two IRs are sufficiently effective for open-source, mid-size LLMs generating PDDL, as one is too distant from the generation target while PDDL itself is too low-resource. We draw inspiration from related work in program synthesis that suggests high-resource languages \citep{cassano2024knowledgetransferhighresourcelowresource,zhang-etal-2025-bridge} may be effect IRs for generating low-resource languages like PDDL. We also note from previous work that the rigid and particular syntax of PDDL calls for a syntax alignment with an IR. Based on these criteria, we additionally propose two fully formal IRs:

\begin{enumerate}
  \setcounter{enumi}{2}
    \item \textbf{Python simulator}, where the LLM is instructed to generate a Python code that simulates the domain and the problem. We place minimal requirement on the structure of the code to leverage the coding ability of high-resource programming languages. We require the generated Python code to be directly executable to return a plan, though in this work treat it as an IR that we transpile to PDDL to take advantage of the optimized planners. 
    \item \textbf{PyPDDL}, where the LLM is instructed to generate Python code in a specific wrapper for PDDL referred to as PyPDDL\footnote{\url{https://github.com/remykarem/Py2PDDL}}. The documentation for PyPDDL is provided in the prompt along with a domain-agnostic example. While the library does not provide a solver, implementing one is feasible as the language is fully formal. Even so, in this work we again transpile the PyPDDL code to PDDL either deterministically with tools from the library or with an LLM.
\end{enumerate}

Examples of IRs are shown in Appendix~\ref{sec:example_ir}.

\paragraph{Choice of Modules}
$I \rightarrow L$: In general planning tasks, there exists much prior work of \lap that generates plans in an end-to-end manner \citep{valmeekam2023planbench,10.5555/3666122.3669442,valmeekam2025a}. This line of work commonly involves LLM \textit{agents}. Common techniques include reasoning before planning \citep{yao2023reactsynergizingreasoningacting}, task decomposition \citep{prasad-etal-2024-adapt}, reflection after planning \citep{majumder2024clin}, continuous learning \citep{wang2024voyager}, and so on.
As a complete survey exists \citep{wei-etal-2025-plangenllms}, we do not dive into more discussions.

$I \rightarrow \text{PDDL}$: In PDDL-based classical planning, most existing work of \laf generates various parts of the PDDL without involving additional IR. The generation target includes the goal states \citep{xie2023translatingnaturallanguageplanning}, problem file \citep{liu2023llm+,zuo-etal-2025-planetarium}, action semantics \citep{zhang-etal-2024-proc2pddl,zhu-etal-2025-language-models,hu-etal-2025-text2world}, domain file \citep{guan2023leveraging,wong2024learning}, and complete PDDL \citep{huang-zhang-2025-limit,gong2025zeroshotiterativeformalizationplanning} like us. Following cited work, we do not consider methods that require supervised training due to the extreme low-resource nature of PDDL. Common inference techniques include basic \textbf{few-shot prompting} (e.g., \citet{xie2023translatingnaturallanguageplanning}), where the LLM is prompted via in-context learning with some exemplars to translate the natural language $D_d$ and/or $D_p$ into PDDL $\mathcal{DF}$ and/or $\mathcal{PF}$. We do this by default with a prompt that introduces the basic structure of PDDL with a domain-agnostic example. Other techniques include sequential generation (e.g., \citet{gong2025zeroshotiterativeformalizationplanning}), where the LLM generates $\mathcal{DF}$ and/or $\mathcal{PF}$ sequentially instead of at once with distinctive prompts, or retrieval of documentation (e.g., \citet{wang2025documentationretrievalimprovesplanning}), where the LLM retrieves the documentation of the PDDL language along with examples before generation. We consider those contemporaneous techniques out of scope.
  
We additionally apply \textbf{decoding by grammar} technique (denoted $\rightarrow^{\text{GD}}$) to PDDL generation based on grammar \citep{10.5555/3666122.3668959} and constrained decoding \citep{scholak-etal-2021-picard} favorable in the program synthesis community. 
We translate the formal BNF definition of PDDL 3.1\footnote{Kovacs, 2011: \url{http://pddl4j.imag.fr/repository/wiki/BNF-PDDL-3.1.pdf}} into a LALR(1)-compatible EBNF grammar used to limit LLMs' decoding to trivially syntactically correct PDDL, but may degrade semantics. 

$\text{PDDL} \leadsto L$: All cited work above use a symbolic PDDL solver such as Fast Downward planner \citep{10.5555/1622559.1622565}, as do we. Another strand of work has used LLMs in lieu of such solvers to take PDDL as input and outputs a plan \citep{stein2025automating}, or to generate programs that emulate such solvers \citep{10.1609/aaai.v38i18.30006,chen2025codeassymbolicplannerfoundationmodelbasedrobot}.

$\text{PDDL} \xrightarrow{}\mathrel{\vphantom{\to}^*}\text{PDDL}$: This represents a revision module, where the LLM first generates an initial PDDL, before revising it multiple times ($\xrightarrow{}\mathrel{\vphantom{\to}^*}$). 
In conjunction with the PDDL solver, the validity of the predicted PDDL along with syntax error messages are typically provided to the LLM for re-generation, denoted as $\rightarrow^{\text{FB}}$ \citep{zhu-etal-2025-language-models,hu-etal-2025-text2world}. 
It is also possible to revise PDDL solely based on LLMs' own feedback without invoking the solver. 

$\text{IR} \xrightarrow{}\mathrel{\vphantom{\to}^*} \text{IR}$: A generated IR can also be revised similarly. However, not all IRs can work with a formal solver, so we rely on the LLM to optionally revise them without indication if their correctness. 

$\text{IR} \rightarrow \text{PDDL}$: Overall, higher-order chaining of general purpose IRs (e.g., natural language, JSON) has proven effective for non-PDDL planning language generation \citep{wong2024learning,hao2025planningrigorgeneralpurposezeroshot}. In our work, we specifically propose IRs with high syntax alignment with PDDL. To transpile an IR to PDDL, we rely on either (i) a deterministic rule-based transpiler or (ii) an LLM. We default to the latter following cited work and discuss the former in a later ablation study.


\paragraph{Non-PDDL Planning Languages}
Even in planning, LLM-as-formalizer is a general methodology that can be instantiated with many planning languages other than PDDL, including satisfiability modulo theories (SMT) \citep{guo2024castlconstraintsspecificationsllm,hao2025planningrigorgeneralpurposezeroshot}, linear temporal logic (LTL) \citep{yang2023plugsafetychipenforcing,li2024embodied}, Answer Set Programming \citep{lin2024clmaspcouplinglargelanguage}, action languages (AL) \citep{ishay2025llmalbridginglargelanguage}, and so on. While our work is largely applicable to any planning language (except for IRs requiring affinity to a particular language), we only study PDDL which is widely used in literature and also the language that underlies many classical planning benchmarks. 

\begin{table*}[t]
    \centering
    \footnotesize
    \newcommand{\new}{\textsuperscript{†}}
    \begin{tabular}{cclllcm{5.8cm}}
        \textbf{Lv.} & \textbf{Input} & \textbf{1st IR} & \textbf{2nd IR} & \textbf{3rd IR} & \textbf{Output} & \textbf{Description} \\ \midrule
        
        0 & $I$ & & & & $\rightarrow L$ & LLM-as-planner ($\phi$) \\ \midrule

        \multirow{3}{*}{1} & $I$ & $\rightarrow \text{PDDL}$ & & & $\leadsto L$ & \textit{PDDL-base}: directly generating PDDL. \\ 
        \cmidrule{2-7}
        & $I$ & $\rightarrow^{\text{GD}} \text{PDDL}$ & & & $\leadsto L$ & \textit{PDDL-grammar}\new the above plus constraining LLM decoding with grammar. \\ 
        \midrule

        \multirow{9}{*}{2} & $I$ & $\rightarrow \text{NL}$ & $\rightarrow \text{PDDL}$ & & $\leadsto L$ & Generating an NL description of the actions and entity states before PDDL \\ 
        \cmidrule{2-7}
        & $I$ & $\rightarrow \text{PyPDDL}$ & $\rightarrow | \leadsto \text{PDDL}$ & & $\leadsto L$ & \new Transpiling a PyPDDL program to PDDL either with Py2PDDL ($\leadsto$) or by an LLM ($\rightarrow$). \\ 
        \cmidrule{2-7}
        & $I$ & $\rightarrow \text{PySim}$ & $\rightarrow \text{PDDL}$ & & $\leadsto L$ & \new Generating a Python-based simulator before translating to PDDL. \\ 
        \cmidrule{2-7}
        & $I$ & $\rightarrow \text{PDDL}$ & $\rightarrow^{\text{FB?}} \text{PDDL}$ & & $\leadsto L$ & Refining the generated PDDL once (with or without the solver feedback). \\ 
        \midrule

        \multirow{9}{*}{3} 
        & $I$ & $\rightarrow \text{PyPDDL}$ & $\rightarrow \text{PDDL}$ & $\rightarrow \text{PDDL}$ & $\leadsto L$ & \new Generating a PyPDDL wrapper before generating PDDL that receives one revision. \\ \cmidrule{2-7}
        & $I$ & $\rightarrow \text{PyPDDL}$ & $\rightarrow \text{PyPDDL}$ & $\rightarrow \text{PDDL}$ & $\leadsto L$ & \new Revising the PyPDDL representation once before generating PDDL. \\
        \cmidrule{2-7}
        & $I$ & $\rightarrow \text{NL}$ & $\rightarrow \text{PyPDDL}$ & $\rightarrow \text{PDDL}$ & $\leadsto L$ & \new Marching through three IRs to emulate a loose-to-formal transpilation process. \\
        \cmidrule{2-7}
        & $I$ & $\rightarrow \text{PDDL}$ & $\rightarrow^{\text{FB?}} \text{PDDL}$ & $\rightarrow^{\text{FB?}} \text{PDDL}$ & $\leadsto L$ & Iteratively refining the PDDL program twice (with or without solver feedback). \\
        
        \bottomrule
    \end{tabular}
    \caption{
        Pipelines that we evaluate in this work, ranging from level 0 to level 3.
        For each pipeline, we specify the IRs before reaching the final plan ($L$).
        Pipelines marked with \new~are newly introduced in this work.
        Straight arrows ($\rightarrow$) denote procedures involving LLMs, with superscripts indicating special variants---GD for grammar-guided decoding and FB for solver feedback.
        Wavy arrows ($\leadsto$) represent purely symbolic solving or compilation processes.
    }
    \label{tab:pipelines}
\end{table*}

\section{Experimental Setup}
\label{data_models_eval}

\noindent \textbf{Datasets.} 
We consider 4 simulated planning environments highly common in cited literature. First, we have BlocksWorld, Logistics, and Sokoban from the International Planning Competition \citep{ipc}. These domains represent object manipulation and range from 4 actions to 12 actions. We use the moderately templated version of the datasets from \citet{huang-zhang-2025-limit}. Next, we have CoinCollector \cite{yuan2019counting}, a partially observable navigation domain, which convert to be fully observable just as the other three. The dataset of each environment comes with 100 tuples of domain description, problem description, ground-truth domain and problem files. Examples are shown in Appendix~\ref{sec:example_input_output}.

\noindent \textbf{Metrics.}
We use syntactic and semantic accuracy to assess the generated PDDL. \textit{Syntactic accuracy} is the percentage where no syntax error are returned by the planning solver.\textit{Plan accuracy} is the percentage where a plan is not only found by the solver but also correct based on validation. We use the \texttt{dual-bfws-ffparser} planner \citep{muise-icaps16demo-pd} to solve for the plan and VAL \citep{1374201} to validate the plan against the ground-truth PDDL. 

\noindent \textbf{Pipelines.}
We instantiate an array of pipelines based on our previously introduced framework. Recall that we define the \textit{level} of a pipeline as the number of IRs (including PDDL) it involves. We notate each pipeline as the intermediate IRs involves, excluding the input description $I$ and the output plan $L$. For example, PDDL $\rightarrow $ PDDL denotes a pipeline $I \rightarrow \text{PDDL} \rightarrow \text{PDDL} \leadsto L$.
The evaluated pipelines are illustrated in Table~\ref{tab:pipelines}, ranging from level 0 to level 3. 
We do not consider level 4 onward pipelines. 


\noindent \textbf{Models.}
We evaluate 4 recent and best performing open-source LLMs, QwQ-32B, Qwen3-32B, gpt-oss-120b (total 117B with 5.1B active parameters), and GLM-4.5-Air-FP8 (total 106B with 12B active parameters)  We use vLLM \citep{kwon2023efficient} to speed up inference and set temperature of 0.4 for each of our experiments on one H100 GPU. We report mean of three runs and their standard deviation is shown as error bars. Examples of prompts are shown in Appendix~\ref{sec:example_prompt}.

\begin{figure*}[t!]
    \centering
    \includegraphics[width=.97\textwidth]{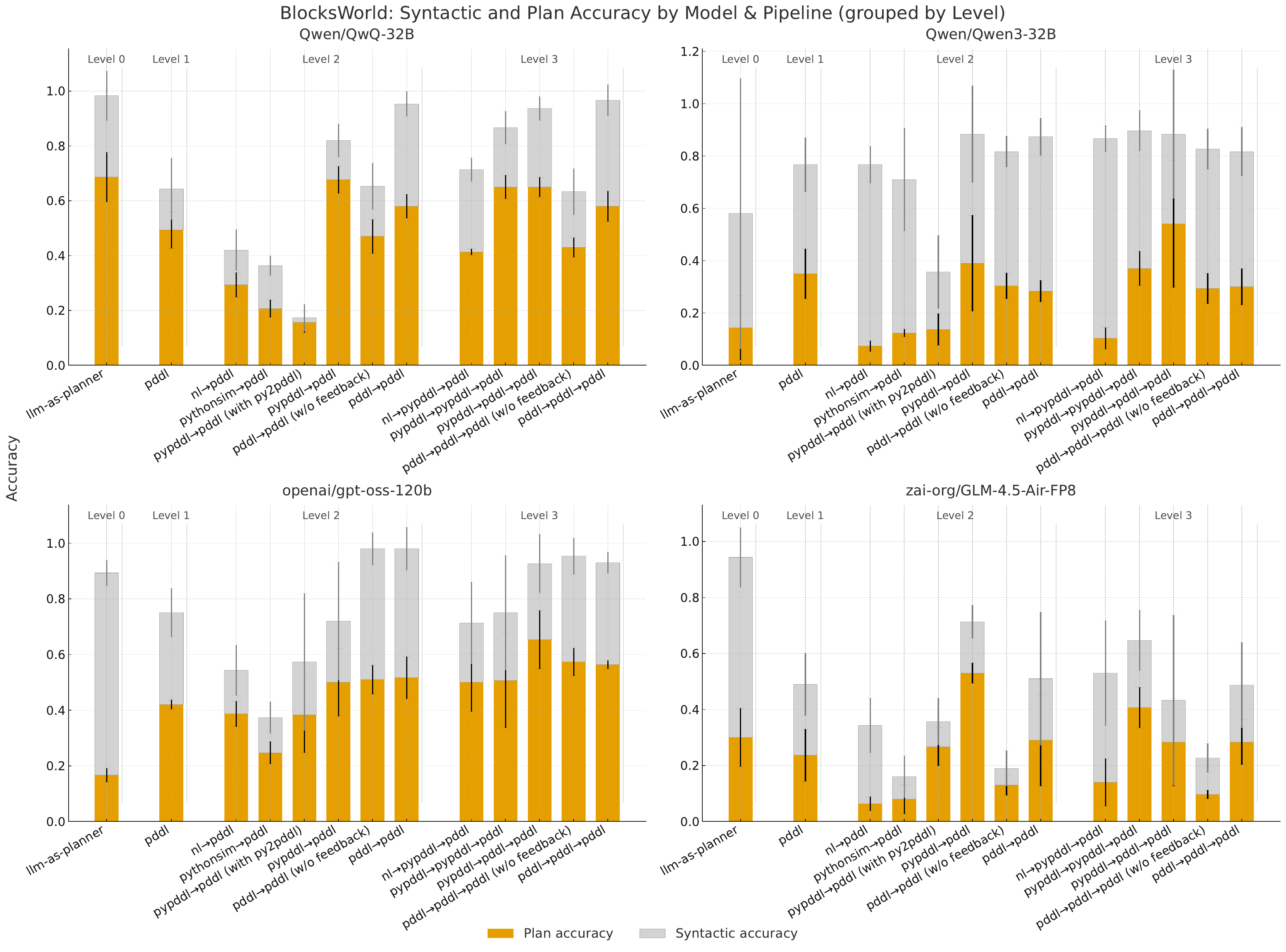}
    \includegraphics[width=.97\textwidth]{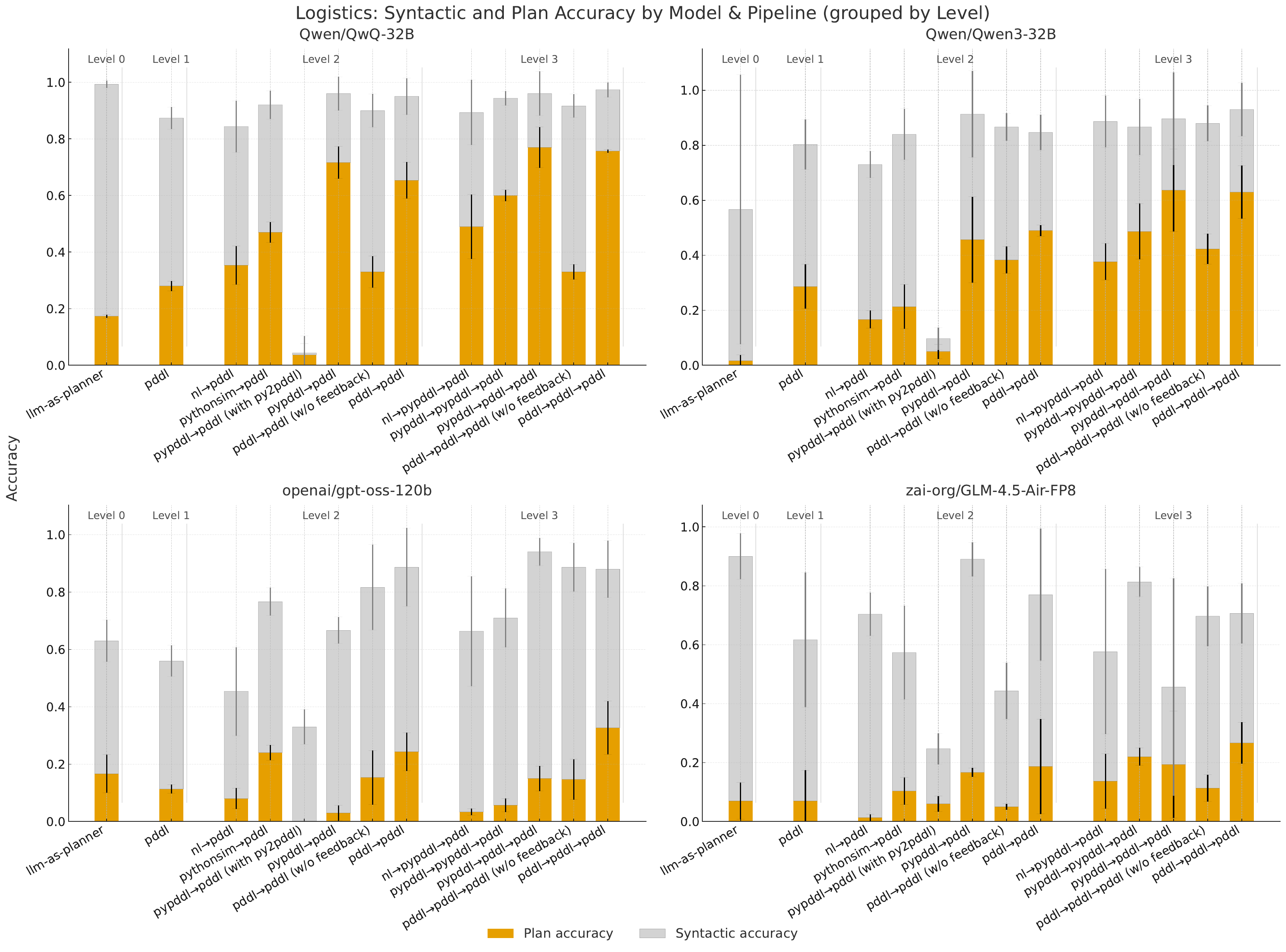}
    \caption{Syntactic accuracy and plan accuracy of various methods grouped by levels and LLMs on BlocksWorld and Logistics. Error bars shows the standard deviation over 3 runs.}
    \label{fig:results}
\end{figure*}

\begin{figure*}[t!]
    \centering
    \includegraphics[width=\textwidth]{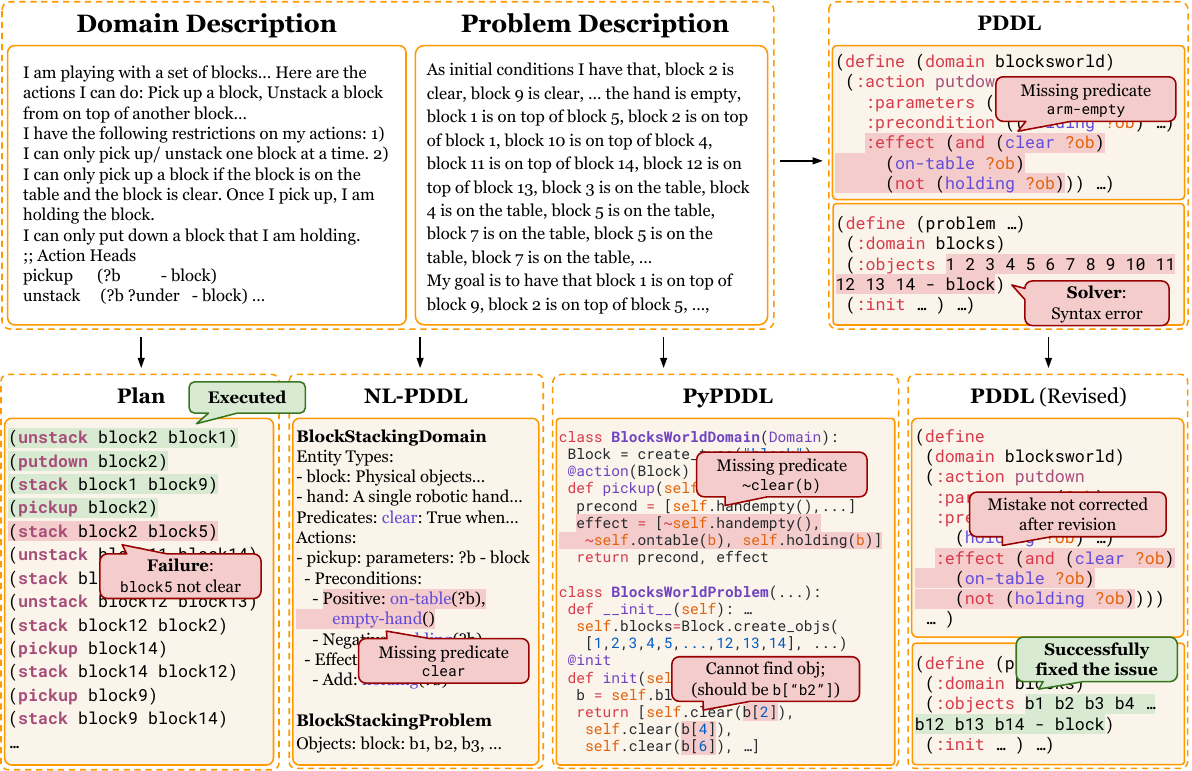}
    \caption{Qualitative examples of a same problem in BlocksWorld, juxtaposing different IRs generated including NL, PyPDDL, PDDL, before generating the final PDDL (Level 2). Also included is a revised PDDL based on solver feedback and a directly generated plan (Level 0).}
    \label{fig:qual_examples}
\end{figure*}

\section{Results and Discussions}

With our unifying framework, we present a comprehensive evaluation for inference-time LLM-assisted planning focusing on PDDL generation. 

\subsection{The Best Pipeline}

Figure~\ref{fig:results} shows the performance of 4 LLMs and 12 pipelines on BlocksWorld and Logistics. The results of Sokoban and CoinCollector are shown in Appendix~\ref{sec:sokoban_cc_results}. We focus on the first two due to space limitation and relatively consistent findings. Overall, the highest plan accuracy is achieved by a Level 3 pipeline, involving 3 IRs, in 6 out of 8 LLM-domain combinations. Among those, pipeline that involves PyPDDL as an IR (PyPDDL → PyPDDL or $\text{PDDL} \rightarrow^{\text{FB}} \text{PDDL}$) performs the best most frequently, while that only involving revisions of PDDL (PyPDDL → PDDL →$^{\text{FB}}$ PDDL) comes as a second. Generally, our finding \textbf{recommends a higher level pipeline that involves IRs and revisions}, for the LLM-as-formalizer methodology. 

However, as real-life planning applications cannot assume unlimited computational resource. Therefore, we now consider the most resource-constrained case of Level 0 and Level 1 pipelines where the most costly step, an LLM generation, happens once. The comparison between the vanilla LLM-as-planner ($\phi$) and LLM-as-formalizer (PDDL-base) has no clear winner (3 against 5 in 8 LLM-domain combinations), corroborating similar claims in recent work that evaluated reasoning LLMs with inference-time scaling \cite{huang-zhang-2025-limit,amonkar2025llmsbetterformalizerssolvers}. 
Applicable for all levels, we evaluate PDDL-grammar with constrained decoding as a drop-in replacement for standard decoding in PDDL-base pipeline. Although performance on BlockWorld domain is comparable (38\% for PDDL-grammar vs 36\% for PDDL-base), the performance on Logistics domain is zero. Constrained decoding showed an hit or miss behaviour; either performing comparable to standard decoding or catastrophically making similar mistakes in all the problems. We defer comprehensive evaluation of constrained decoding to future work while suggesting for robust decoding techniques.
With one more IR, \textbf{the best Level 2 LLM-as-formalizer pipeline always outperforms LLM-as-planner} and also the Level 1 pipelines in all 8 out of 8 cases. This finding cautions against generating planning languages with merely one LLM call. 

\begin{figure*}[t!]
    \centering
    \includegraphics[width=\textwidth]{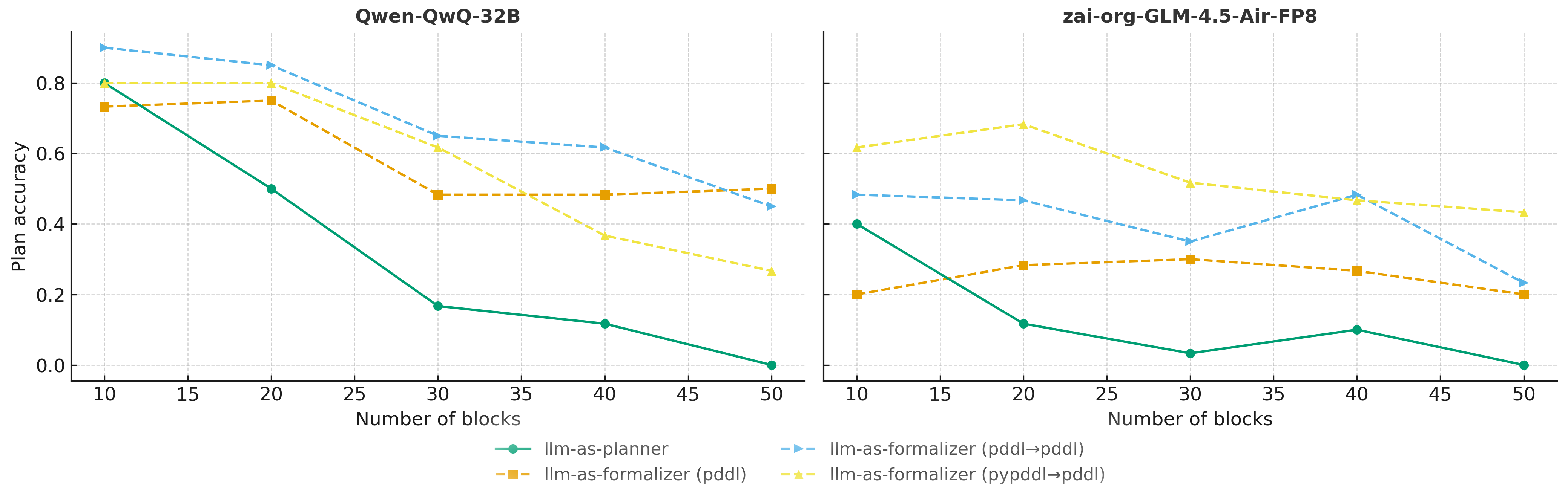}
    \caption{The performance of two LLMs and four methods including usage as planners and as formalizers on BlocksWorld problems with an increasing entity space. }
    \label{fig:complexity_analysis}
\end{figure*}

\subsection{The Best IR}
Having demonstrated the effectiveness of IR in planning language generation and justified our IR-based framework, we zoom onto Level 2 with one additional IR before the final PDDL to be generated. 
Our pipelines are instantiated with 4 types of IR: NL, Python Simulator, PyPDDL, and PDDL. Compared to Level 1, NL as an IR consistently hurts performance. Python Simulator is similar yet sometimes improves performance. In contrast, \textbf{PyPDDL and PDDL as IRs consistently improves performance}. 
Notably, PyPDDL → PDDL and $\text{PDDL} \rightarrow^{\text{FB}} \text{PDDL}$ out-performs each other in half of the 8 LLM-domain combinations. Without the solver feedback but with the LLM feedback, the PDDL → PDDL pipeline loses its advantage of revision. 
Whenever invoking the solver may be costly, this finding presents the community an alternative to revision with solver feedback: transpilation from a syntax-aligned IR to the target language. 
Figure~\ref{fig:qual_examples} shows qualitative examples of different IRs and their typical mistakes.

For any pipeline that involves PyPDDL → PDDL, recall that PyPDDL is a formal language where a solver \textit{could} be defined in situ (though, this is a non-trivial effort) yet is not provided by the library we use, so we default to using the LLM to transpile PyPDDL to PDDL to leverage the PDDL solver. As an ablation study, we also attempt to perform transpilation deterministically using Py2PDDL library (denoted as $\text{PyPDDL} \leadsto \text{PDDL}$). 
One may expect such a deterministic transpilation to outperform LLM generating PDDL, considering Python is much higher-resource than PDDL. 
Transpilation with Py2PDDL performs much worse than LLM-based transpilation. 
For instance, with accuracy 16\% vs 68\% for QwQ-32B model in BlocksWorld. 
The generated PyPDDL tends to not follow non-trivial way of representing different PDDL components. 
As shown in Figure~\ref{fig:qual_examples}, a set of \texttt{blocks} created by \texttt{create\_objs} is mistakenly treated as a \texttt{List} instead of a \texttt{Dict}.






\subsection{Robustness against Complexity}

Recent work has cast doubt on the ability of LLM, both as a planner and as a formalizer, to scale with the problem complexity \cite{shojaee2025illusionthinkingunderstandingstrengths,lin2025zebralogic,amonkar2025llmsbetterformalizerssolvers}. For example, the BlocksWorld dataset evaluated in this work ranges from 2 to 15 blocks, while other cited previous work often relied on datasets with even smaller entity space. In real-life applications such as robotics or logistics, current benchmarks do not help understand LLMs' robustness in planning for complex problems, either as planners or as formalizers. Therefore, we specifically construct BlocksWorld problems with increasing number of entities (blocks), ranging from 10 to 50, with an increment of 10 and 10 problems per bucket. We evaluate the robustness of some of the most commonly used and best-performing pipelines LLM-as-planner (Level 0) and three LLM-as-formalizer pipelines including PDDL (Level 1), PyPDDL → PDDL, and PDDL → PDDL (Level 2).

Figure~\ref{fig:complexity_analysis}, clearly shows that while \textbf{all methods degrade with complexity, LLM-as-formalizer is much more robust than LLM-as-planner}. We observe that, as the number of blocks approaches 20, the performance of LLM-as-planner halves compared to 10; approaching 50, its performance drops to zero. In contrast, all LLM-as-formalizer pipelines are more robust to increased entity-space complexity despite reduction in performance. Notably, directly generating PDDL (Level 1) loses no more than 20\% plan accuracy as the number of entities increases from 10 to 50. In comparison, the Level 2 pipelines perform better with less entities, but degrade more severely with more entities. 

Situated among recent studies, our findings corroborate the claim that even reasoning LLMs with inference-time scaling are not robust to problem complexity. Moreover, we are the first to evidence and advocate for the robustness of LLM-as-formalizer in planning. For more multi-IR pipelines of higher levels, we also caution against the tradeoff between performance and robustness against complexity.

\section{Conclusion}
We propose a unifying framework of inference-time planning language generation. Based on the interplay of IR, our framework not only subsumes most prior work involving multi-stage LLM pipelines but also inspires novel pipelines, such as those involving a high-resource, syntactically similar IR to the target language. In addition to providing immediately applicable recipes for high-performing planning methods, our work opens up promising directions of research in designing efficient LLM pipelines that outreaches to alternative planning languages (e.g., LTL) and even non-planning domain-specific languages (e.g., SMT).

\section{Limitations}

We work with moderately templated descriptions for domain and problem as input. This is easier than more natural version of the data. While we observe increased performance with some of the techniques implemented, it is not clear whether we same performance gains with natural data which is more applicable to the real world.

We only consider pipelines with three levels i.e. three IRs for planning language generation. Some of the existing work already have more than three IRs in their pipelines \citep{hao2025planningrigorgeneralpurposezeroshot, ishay2025llmalbridginglargelanguage}. Thus, instantiation of our framework presented here is still limited in all possible insights applicable to contemporary work. We defer this to future work.



\bibliography{anthology, custom}

\appendix

\section{Additional Results}
\label{sec:sokoban_cc_results}
\begin{figure*}[t!]
    \centering
    \includegraphics[width=.95\textwidth]{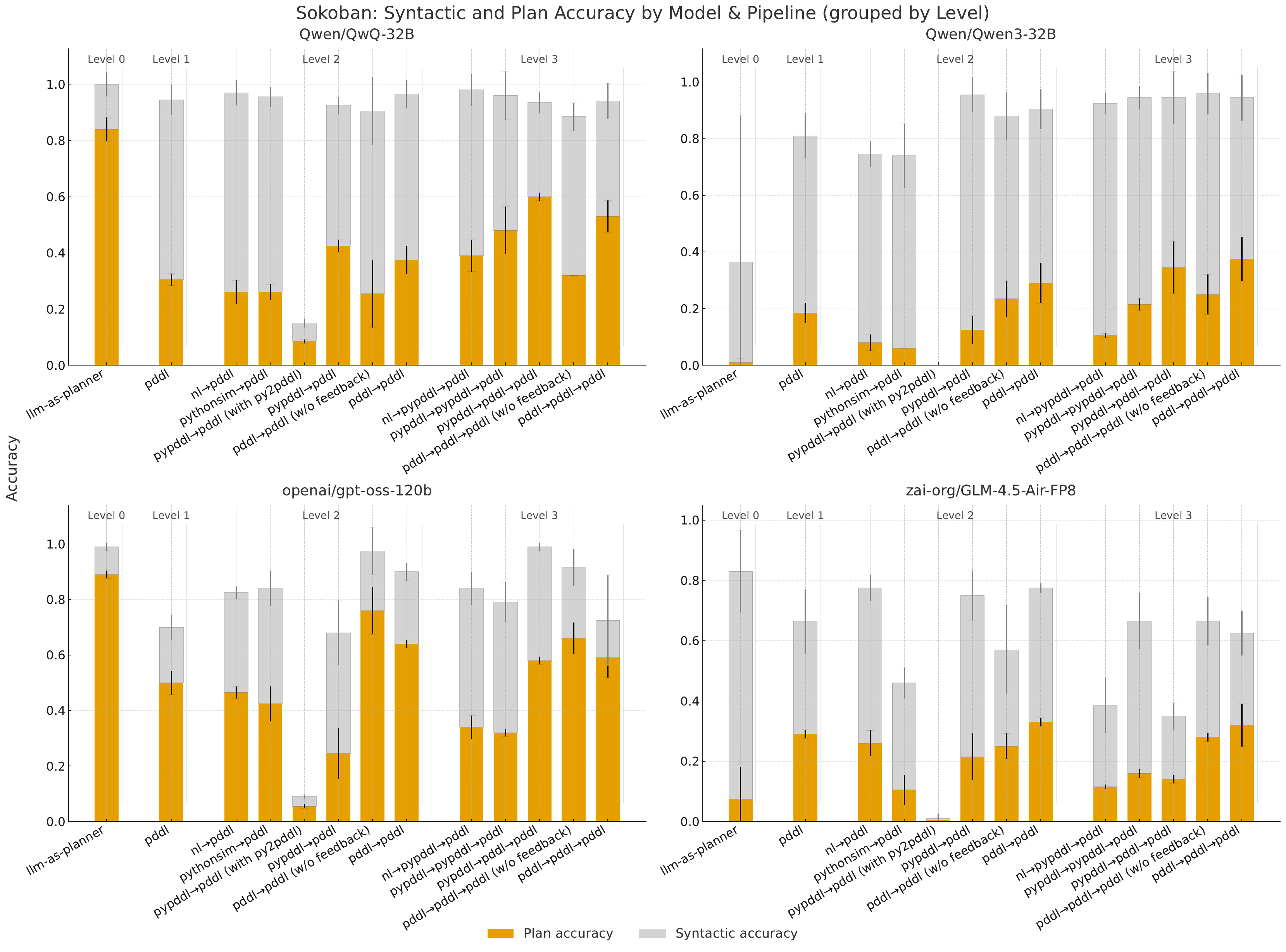}
    \includegraphics[width=.95\textwidth]{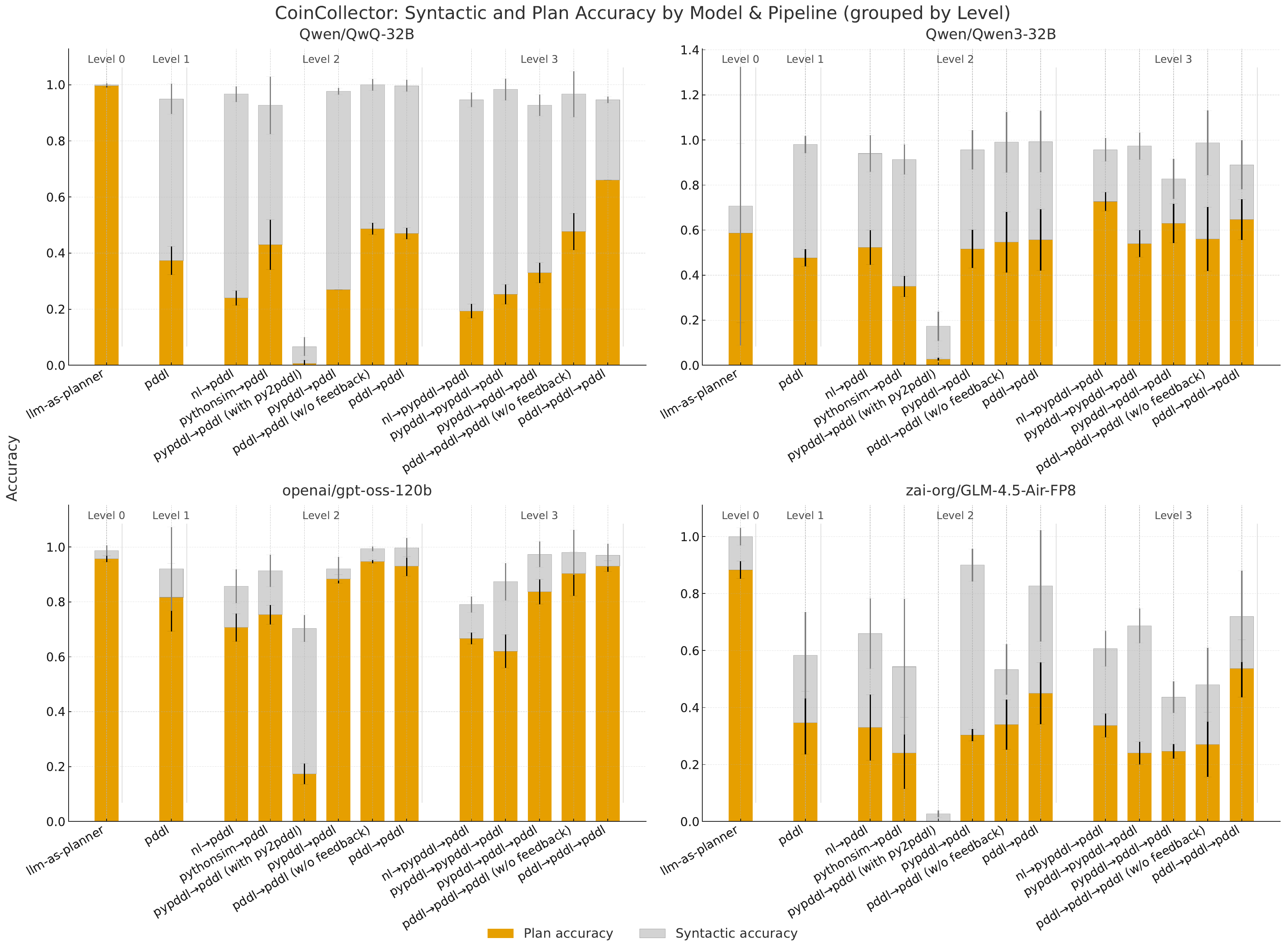}
    \caption{Syntactic accuracy and plan accuracy of various methods grouped by levels and LLMs on Sokoban and CoinCollector. Error bars shows the standard deviation over 3 runs.}
    \label{fig:sokoban_cc_results}
\end{figure*}

The experimental results for Sokoban and CoinCollector are shown in Table~\ref{fig:sokoban_cc_results}. 

\section{Examples of Input and Output}
\label{sec:example_input_output}

\subsection{BlocksWorld}
\paragraph{Blocksworld.} Find the domain description (dd; Listing~\ref{lst:blocksworld-dd}), problem description (pd; Listing~\ref{lst:blocksworld-pd}), domain file (df; Listing~\ref{lst:blocksworld-df}), problem file (pf; Listing~\ref{lst:blocksworld-pf}), and plan (plan; Listing~\ref{lst:blocksworld-plan}).

\begin{figure*}[t]
  \centering
  \begin{minipage}{\textwidth}
\begin{lstlisting}[language=,basicstyle=\ttfamily\small]
I am playing with a set of blocks where I need to arrange the blocks into stacks. Here are the actions I can do

   Pick up a block
   Unstack a block from on top of another block
   Put down a block
   Stack a block on top of another block

   I have the following restrictions on my actions:
   I can only pick up or unstack one block at a time.
   I can only pick up or unstack a block if my hand is empty.
   I can only pick up a block if the block is on the table and the block is clear. A block is clear if the block has no other blocks on top of it and if the block is not picked up.
   I can only unstack a block from on top of another block if the block I am unstacking was really on top of the other block.
   I can only unstack a block from on top of another block if the block I am unstacking is clear.
   Once I pick up or unstack a block, I am holding the block.
   I can only put down a block that I am holding.
   I can only stack a block on top of another block if I am holding the block being stacked.
   I can only stack a block on top of another block if the block onto which I am stacking the block is clear.
   Once I put down or stack a block, my hand becomes empty.
   Once you stack a block on top of a second block, the second block is no longer clear.

;; Action Heads
pickup      (?b          - block)
unstack     (?b ?under   - block)
putdown     (?b          - block)
stack       (?b ?under   - block)
\end{lstlisting}
  \end{minipage}
  \caption{Blocksworld Domain Description}
  \label{lst:blocksworld-dd}
\end{figure*}

\begin{figure*}[t]
  \centering
  \begin{minipage}{\textwidth}
\begin{lstlisting}[language=,basicstyle=\ttfamily\small]
As initial conditions I have that, block1 is clear, block2 is clear, block3 is clear, block5 is clear, the hand is empty, block5 is on top of block4, block1 is on the table, block2 is on the table, block3 is on the table, and block4 is on the table.
My goal is to have that block4 is on top of block5, block1 is on the table, block2 is on the table, block3 is on the table, and block5 is on the table.
\end{lstlisting}
  \end{minipage}
  \caption{Blocksworld Problem Description}
  \label{lst:blocksworld-pd}
\end{figure*}

\begin{figure*}[t]
  \centering
  \begin{minipage}{\textwidth}
\begin{lstlisting}[language=,basicstyle=\ttfamily\small]
(define (domain blocksworld)
  (:requirements :strips)
(:predicates (clear ?x)
             (on-table ?x)
             (arm-empty)
             (holding ?x)
             (on ?x ?y))

(:action pickup
  :parameters (?ob)
  :precondition (and (clear ?ob) (on-table ?ob) (arm-empty))
  :effect (and (holding ?ob) (not (clear ?ob)) (not (on-table ?ob)) 
               (not (arm-empty))))

(:action putdown
  :parameters  (?ob)
  :precondition (holding ?ob)
  :effect (and (clear ?ob) (arm-empty) (on-table ?ob) 
               (not (holding ?ob))))

(:action stack
  :parameters  (?ob ?underob)
  :precondition (and (clear ?underob) (holding ?ob))
  :effect (and (arm-empty) (clear ?ob) (on ?ob ?underob)
               (not (clear ?underob)) (not (holding ?ob))))

(:action unstack
  :parameters  (?ob ?underob)
  :precondition (and (on ?ob ?underob) (clear ?ob) (arm-empty))
  :effect (and (holding ?ob) (clear ?underob)
               (not (on ?ob ?underob)) (not (clear ?ob)) (not (arm-empty)))))
\end{lstlisting}
  \end{minipage}
  \caption{Blocksworld Domain PDDL}
  \label{lst:blocksworld-df}
\end{figure*}

\begin{figure*}[t]
  \centering
  \begin{minipage}{\textwidth}
\begin{lstlisting}[language=,basicstyle=\ttfamily\small]
(define (problem blocksworld-p02)
  (:domain blocksworld)
  (:objects block1 block2 block3 block4 block5 )
  (:init 
    (on-table block1)
    (clear block1)
    (on-table block3)
    (clear block3)
    (on-table block2)
    (clear block2)
    (on-table block4)
    (on block5 block4)
    (clear block5)
    (arm-empty)
  )
  (:goal (and 
    (on-table block3)
    (on-table block5)
    (on block4 block5)
    (on-table block2)
    (on-table block1)
  ))
)
\end{lstlisting}
  \end{minipage}
  \caption{Blocksworld Problem PDDL}
  \label{lst:blocksworld-pf}
\end{figure*}

\begin{figure*}[t]
  \centering
  \begin{minipage}{\textwidth}
\begin{lstlisting}[language=,basicstyle=\ttfamily\small]
1. (unstack block5 block4)
2. (putdown block5)
3. (pickup block4)
4. (stack block4 block5)
\end{lstlisting}
  \end{minipage}
  \caption{Blocksworld Plan}
  \label{lst:blocksworld-plan}
\end{figure*}

\subsection{Logistics}
\paragraph{Logistics.} Find the domain description (dd; Listing~\ref{lst:logistics-dd}), problem description (pd; Listing~\ref{lst:logistics-pd}), domain file (df; Listings~\ref{lst:logistics-df-a} and \ref{lst:logistics-df-b}), problem file (pf; Listing~\ref{lst:logistics-pf}), and plan (plan; Listing~\ref{lst:logistics-plan}).

\begin{figure*}[t]
  \centering
  \begin{minipage}{\textwidth}
\begin{lstlisting}[language=,basicstyle=\ttfamily\small]
I need to move packages between locations. Here are the actions I can do

    Load an package onto a truck at a location (load-truck package truck location)
    Load an package onto an airplane at a location (load-airplane package airplane location)
    Unload an package from a truck at a location (unload-truck package truck location)
    Unload an package from an airplane at a location (unload-airplane package airplane location)
    Drive a truck from location1 to location2 in a city (drive-truck truck location1 location2 city)
    Fly an airplane from airport1 to airport2 (fly-airplane airplane airport1 airport2)
    
I have the following restrictions on my actions:
    I can only load a package onto a truck or airplane if both the package and airplane are at the location.
    Once I load the package in the truck or airplane, it is no longer at the location.
    I can only unload a package from a truck or airplane if the truck or airplane is at the location and the package is in the truck or airplane.
    Once I unload the truck or airplane, the object is at the location and no longer in the truck or airplane.
    I can only drive a truck between locations if the truck is at the first location and both the first and second locations are in the same city. Once I drive a truck, the truck is in the second city and no longer in the first city.
    I can only fly an airplane between two airports and the airplane is at the first airport.
    Once I fly an airplane, the airplane is at the second airport and no longer at the first airport.

;; Action Heads
load-truck        (?obj - package   ?truck - truck     ?loc - location)
load-airplane     (?obj - package   ?airplane - airplane  ?loc - airport)
unload-truck      (?obj - package   ?truck - truck     ?loc - location)
unload-airplane   (?obj - package   ?airplane - airplane  ?loc - airport)
drive-truck       (?truck - truck   ?loc-from - location  ?loc-to - location  ?city - city)
fly-airplane      (?airplane - airplane  ?loc-from - airport  ?loc-to - airport)
\end{lstlisting}
  \end{minipage}
  \caption{Logistics Domain Description}
  \label{lst:logistics-dd}
\end{figure*}

\begin{figure*}[t]
  \centering
  \begin{minipage}{\textwidth}
\begin{lstlisting}[language=,basicstyle=\ttfamily\small]
As initial conditions, I have that, obj11 is a package, obj12 is a package, obj13 is a package, obj21 is a package, obj22 is a package, obj23 is a package, tru1 is a truck, tru2 is a truck, cit1 is a city, cit2 is a city, pos1 is a location, apt1 is a location, pos2 is a location, apt2 is a location, apt1 is an airport, apt2 is an airport, apn1 is an airplane, apn1 is at apt2, tru1 is at pos1, obj11 is at pos1, obj12 is at pos1, obj13 is at pos1, tru2 is at pos2, obj21 is at pos2, obj22 is at pos2, obj23 is at pos2, pos1 is in cit1, apt1 is in cit1, pos2 is in cit2, and apt2 is in cit2. 
My goal is to have that obj12 is at pos1, obj23 is at pos1, obj11 is at apt1, obj22 is at apt1, obj13 is at pos2, and obj21 is at pos2. 
\end{lstlisting}
  \end{minipage}
  \caption{Logistics Problem Description}
  \label{lst:logistics-pd}
\end{figure*}

\begin{figure*}[t]
  \centering
  \begin{minipage}{\textwidth}
\begin{lstlisting}[language=,basicstyle=\ttfamily\small]
;; logistics domain
;;

(define (domain logistics)
  (:requirements :strips) 
  (:predicates 	(package ?obj)
	       (truck ?truck)
		(airplane ?airplane)
                (airport ?airport)
                	(location ?loc)
		(in-city ?obj ?city)
                (city ?city)
		(at ?obj ?loc)
		(in ?obj ?obj))

 
(:action load-truck
  :parameters
   (?obj
    ?truck
    ?loc)
  :precondition
   (and (package ?obj) (truck ?truck) (location ?loc)
   (at ?truck ?loc) (at ?obj ?loc))
  :effect
   (and (not (at ?obj ?loc)) (in ?obj ?truck)))

(:action load-airplane
  :parameters
   (?obj
    ?airplane
    ?loc)
  :precondition
   (and (package ?obj) (airplane ?airplane) (location ?loc)
   (at ?obj ?loc) (at ?airplane ?loc))
  :effect
   (and (not (at ?obj ?loc)) (in ?obj ?airplane)))

(:action unload-truck
  :parameters
   (?obj
    ?truck
    ?loc)
  :precondition
   (and (package ?obj) (truck ?truck) (location ?loc)
        (at ?truck ?loc) (in ?obj ?truck))
  :effect
   (and (not (in ?obj ?truck)) (at ?obj ?loc)))

(:action unload-airplane
  :parameters
   (?obj
    ?airplane
    ?loc)
  :precondition
   (and (package ?obj) (airplane ?airplane) (location ?loc)
        (in ?obj ?airplane) (at ?airplane ?loc))
  :effect
   (and (not (in ?obj ?airplane)) (at ?obj ?loc)))
\end{lstlisting}
  \end{minipage}
  \caption{Logistics Domain PDDL (Part 1)}
  \label{lst:logistics-df-a}
\end{figure*}

\begin{figure*}[t]
  \centering
  \begin{minipage}{\textwidth}
\begin{lstlisting}[language=,basicstyle=\ttfamily\small]
; Logistics domain (continued)

(:action drive-truck
  :parameters
   (?truck
    ?loc-from
    ?loc-to
    ?city)
  :precondition
   (and (truck ?truck) (location ?loc-from) (location ?loc-to) (city ?city)
   (at ?truck ?loc-from)
   (in-city ?loc-from ?city)
   (in-city ?loc-to ?city))
  :effect
   (and (not (at ?truck ?loc-from)) (at ?truck ?loc-to)))

(:action fly-airplane
  :parameters
   (?airplane
    ?loc-from
    ?loc-to)
  :precondition
   (and (airplane ?airplane) (airport ?loc-from) (airport ?loc-to)
	(at ?airplane ?loc-from))
  :effect
   (and (not (at ?airplane ?loc-from)) (at ?airplane ?loc-to)))
)
\end{lstlisting}
  \end{minipage}
  \caption{Logistics Domain PDDL (Part 2)}
  \label{lst:logistics-df-b}
\end{figure*}

\begin{figure*}[t]
  \centering
  \begin{minipage}{\textwidth}
\begin{lstlisting}[language=,basicstyle=\ttfamily\small]
(define (problem logistics-6-9)
(:domain logistics)
(:objects apn1 apt2 pos2 apt1 pos1 cit2 cit1 tru2 tru1 obj23 obj22 obj21 obj13 obj12 obj11 )
(:init (package obj11) (package obj12) (package obj13) (package obj21)
 (package obj22) (package obj23) (truck tru1) (truck tru2) (city cit1) (city cit2)
 (location pos1) (location apt1) (location pos2) (location apt2) (airport apt1)
 (airport apt2) (airplane apn1) (at apn1 apt2) (at tru1 pos1) (at obj11 pos1)
 (at obj12 pos1) (at obj13 pos1) (at tru2 pos2) (at obj21 pos2) (at obj22 pos2)
 (at obj23 pos2) (in-city pos1 cit1) (in-city apt1 cit1) (in-city pos2 cit2)
 (in-city apt2 cit2))
(:goal (and (at obj12 pos1) (at obj23 pos1) (at obj11 apt1) (at obj22 apt1)
            (at obj13 pos2) (at obj21 pos2)))
)
\end{lstlisting}
  \end{minipage}
  \caption{Logistics Problem PDDL}
  \label{lst:logistics-pf}
\end{figure*}

\begin{figure*}[t]
  \centering
  \begin{minipage}{\textwidth}
\begin{lstlisting}[language=,basicstyle=\ttfamily\small]
1. (load-truck obj11 tru1 pos1)
2. (load-truck obj13 tru1 pos1)
3. (drive-truck tru1 pos1 apt1 cit1)
4. (unload-truck obj11 tru1 apt1)
5. (unload-truck obj13 tru1 apt1)
6. (load-truck obj22 tru2 pos2)
7. (load-truck obj23 tru2 pos2)
8. (drive-truck tru2 pos2 apt2 cit2)
9. (unload-truck obj22 tru2 apt2)
10. (unload-truck obj23 tru2 apt2)
11. (load-airplane obj22 apn1 apt2)
12. (load-airplane obj23 apn1 apt2)
13. (fly-airplane apn1 apt2 apt1)
14. (unload-airplane obj22 apn1 apt1)
15. (unload-airplane obj23 apn1 apt1)
16. (load-airplane obj13 apn1 apt1)
17. (fly-airplane apn1 apt1 apt2)
18. (unload-airplane obj13 apn1 apt2)
19. (load-truck obj13 tru2 apt2)
20. (drive-truck tru2 apt2 pos2 cit2)
21. (unload-truck obj13 tru2 pos2)
22. (load-truck obj23 tru1 apt1)
23. (drive-truck tru1 apt1 pos1 cit1)
24. (unload-truck obj23 tru1 pos1)
\end{lstlisting}
  \end{minipage}
  \caption{Logistics Plan}
  \label{lst:logistics-plan}
\end{figure*}

\subsection{CoinCollector}

\paragraph{CoinCollector.} Find the domain description (dd; Listing~\ref{lst:coincollector-dd}), problem description (pd; Listing~\ref{lst:coincollector-pd}), domain file (df; Listing~\ref{lst:coincollector-df}), problem file (pf; Listing~\ref{lst:coincollector-pf}), and plan (plan; Listing~\ref{lst:coincollector-plan}).

\begin{figure*}[t]
  \centering
  \begin{minipage}{\textwidth}
\begin{lstlisting}[language=,basicstyle=\ttfamily\small]
I'm exploring a grid of rooms to collect items. Doors may block passage between rooms until I open them. Here are the actions I can do:

    Move from one room to an adjacent room in a given direction
    Open a closed door between two rooms (which then allows travel both ways)
    Take (pick up) an item that's in the same room as me

I have the following restrictions on my actions:

    I can move only if there is an open connection from my current room to the target room in the intended direction.
    If a door between two rooms is closed, I must open it before I can move through it.
    To open a door, I must be in one of the two rooms the door connects, and I must know both the direction I'm opening in and its reverse (e.g., east/west, north/south).
    Opening the door removes the "closed" state and creates connections in both directions.
    I can only take an item if I am in the same room as the item and the item hasn't already been taken.
    Once I take an item, it is considered collected (no longer present in the room).

;; Action Heads

move  (?room1 - room ?room2 - room ?direction - direction)
open  (?room1 - room ?room2 - room ?direction - direction ?reverse - direction)
take  (?item  - item ?room  - room)
\end{lstlisting}
  \end{minipage}
  \caption{CoinCollector Domain Description}
  \label{lst:coincollector-dd}
\end{figure*}

\begin{figure*}[t]
  \centering
  \begin{minipage}{\textwidth}
\begin{lstlisting}[language=,basicstyle=\ttfamily\small]
As initial conditions, I have that, kitchen is a room, corridor is a room, pantry is a room, north is a direction, south is a direction, east is a direction, west is a direction, coin is an item, I am at kitchen, there is an open connection from kitchen to corridor going north, there is an open connection from corridor to kitchen going south, there is a closed door from kitchen to pantry going east, there is a closed door from pantry to kitchen going west, coin is in corridor, direction north is reverse of south, direction south is reverse of north, direction east is reverse of west, and direction west is reverse of east.
My goal is to have that coin is taken.
\end{lstlisting}
  \end{minipage}
  \caption{CoinCollector Problem Description}
  \label{lst:coincollector-pd}
\end{figure*}

\begin{figure*}[t]
  \centering
  \begin{minipage}{\textwidth}
\begin{lstlisting}[language=,basicstyle=\ttfamily\small]
(define (domain coin-collector)
  (:requirements :strips :typing)
  (:types
    room
    direction
    item
  )
  (:predicates
    (at ?room - room)
    (connected ?room1 - room ?room2 - room ?direction - direction)
    (closed-door ?room1 - room ?room2 - room ?direction - direction)
    (location ?item - item ?room - room)
    (taken ?item - item)
    (is-reverse ?direction - direction ?reverse - direction)
  )

  (:action move
    :parameters (?room1 - room ?room2 - room ?direction - direction)
    :precondition (and (at ?room1) (connected ?room1 ?room2 ?direction))
    :effect (and (not (at ?room1)) (at ?room2))
  )

  (:action open
    :parameters (?room1 - room ?room2 - room ?direction - direction ?reverse - direction)
    :precondition (and (at ?room1) (closed-door ?room1 ?room2 ?direction) (is-reverse ?direction ?reverse))
    :effect (and (not (closed-door ?room1 ?room2 ?direction)) (not (closed-door ?room2 ?room1 ?reverse)) (connected ?room1 ?room2 ?direction) (connected ?room2 ?room1 ?reverse))
  )
  
  (:action take
    :parameters (?item - item ?room - room)
    :precondition (and (at ?room) (location ?item ?room) (not (taken ?item)))
    :effect (and (taken ?item) (not (location ?item ?room)))
  )
)
\end{lstlisting}
  \end{minipage}
  \caption{CoinCollector Domain PDDL}
  \label{lst:coincollector-df}
\end{figure*}

\begin{figure*}[t]
  \centering
  \begin{minipage}{\textwidth}
\begin{lstlisting}[language=,basicstyle=\ttfamily\small]
(define (problem coin_collector_numLocations3_numDistractorItems0_seed54)
  (:domain coin-collector)
  (:objects
    kitchen corridor pantry - room
    north south east west - direction
    coin - item
  )
  (:init
    (at kitchen)
    (connected kitchen corridor north)
    (closed-door kitchen pantry east)
    (connected corridor kitchen south)
    (closed-door pantry kitchen west)
    (location coin corridor)
    (is-reverse north south)
    (is-reverse south north)
    (is-reverse east west)
    (is-reverse west east)
  )
  (:goal 
    (taken coin)
  )
)
\end{lstlisting}
  \end{minipage}
  \caption{CoinCollector Problem PDDL}
  \label{lst:coincollector-pf}
\end{figure*}

\begin{figure*}[t]
  \centering
  \begin{minipage}{\textwidth}
\begin{lstlisting}[language=,basicstyle=\ttfamily\small]
1. (move kitchen corridor north)
2. (take coin corridor)
\end{lstlisting}
  \end{minipage}
  \caption{CoinCollector Plan}
  \label{lst:coincollector-plan}
\end{figure*}

\subsection{Sokoban}

\paragraph{Sokoban.} Find the domain description (dd; Listing~\ref{lst:sokoban-dd}), problem description (pd; Listing~\ref{lst:sokoban-pd}), domain file (df; Listings~\ref{lst:sokoban-df-a} and \ref{lst:sokoban-df-b}), problem file (pf; Listing~\ref{lst:sokoban-pf}), and plan (plan; Listing~\ref{lst:sokoban-plan}).

\begin{figure*}[t]
  \centering
  \begin{minipage}{\textwidth}
\begin{lstlisting}[language=,basicstyle=\ttfamily\small]
I am playing a Sokoban puzzle where a player pushes stone crates onto goal squares. Here are the actions I can do

   Move the player into an adjacent clear square
   Push a stone into an adjacent goal square
   Push a stone into an adjacent non-goal square

   I have the following restrictions on my actions:
   I can only move if I choose one of the four cardinal directions, I am the player, I am at the from-square, the destination square is clear, and the two squares are aligned with that direction.
   When I move, I leave my previous square clear and occupy the destination square.
   I can only push a stone toward a goal square if I am the player, the object is a stone, I stand directly behind the stone in the chosen direction, the target square is clear, that target square is marked as a goal, and both adjacency relations match the direction of push.
   After pushing a stone toward a goal square, I advance into the stone's former square, the stone moves into the goal square, the goal square is no longer clear, my previous square becomes clear, and the stone is marked as being on a goal square.
   I can only push a stone toward a non-goal square if I am the player, the object is a stone, I stand behind the stone in the direction of push, the destination square is clear, that destination square is marked as a normal square, and the adjacency relations match the direction of push.
   After pushing a stone toward a non-goal square, I move into the stone's former square, the stone shifts forward, the destination square is no longer clear, and my previous square becomes clear while the stone is marked as not being on a goal square.

;; Action Heads
move             (?p - thing ?from - location ?to - location ?dir - direction)
push-to-goal     (?p - thing ?s - thing ?ppos - location ?from - location ?to - location ?dir - direction)
push-to-nongoal  (?p - thing ?s - thing ?ppos - location ?from - location ?to - location ?dir - direction)
\end{lstlisting}
  \end{minipage}
  \caption{Sokoban Domain Description}
  \label{lst:sokoban-dd}
\end{figure*}

\begin{figure*}[t]
  \centering
  \begin{minipage}{\textwidth}
\begin{lstlisting}[language=,basicstyle=\ttfamily\small]
As initial conditions I have that, player-01 is at pos-1-1, stone-01 is at pos-1-2, pos-1-3 is clear, pos-1-3 is a goal square, pos-1-1 is a normal square, pos-1-2 is a normal square, player-01 is the player, stone-01 is a stone, dir-right is an available move direction, dir-left is an available move direction, dir-up is an available move direction, dir-down is an available move direction, moving from pos-1-1 to pos-1-2 uses dir-right, moving from pos-1-2 to pos-1-1 uses dir-left, moving from pos-1-2 to pos-1-3 uses dir-right, and moving from pos-1-3 to pos-1-2 uses dir-left.
My goal is to have that stone-01 is at a goal square.
\end{lstlisting}
  \end{minipage}
  \caption{Sokoban Problem Description}
  \label{lst:sokoban-pd}
\end{figure*}

\begin{figure*}[t]
  \centering
  \begin{minipage}{\textwidth}
\begin{lstlisting}[language=,basicstyle=\ttfamily\small]
(define (domain sokoban)
  (:requirements :typing )
  (:types thing location direction)
  (:predicates (move-dir ?v0 - location ?v1 - location ?v2 - direction)
	(is-nongoal ?v0 - location)
	(clear ?v0 - location)
	(is-stone ?v0 - thing)
	(at ?v0 - thing ?v1 - location)
	(is-player ?v0 - thing)
	(at-goal ?v0 - thing)
	(move ?v0 - direction)
	(is-goal ?v0 - location)
  )

  ; (:actions move)

  

	(:action move
		:parameters (?p - thing ?from - location ?to - location ?dir - direction)
		:precondition (and (move ?dir)
			(is-player ?p)
			(at ?p ?from)
			(clear ?to)
			(move-dir ?from ?to ?dir))
		:effect (and
			(not (at ?p ?from))
			(not (clear ?to))
			(at ?p ?to)
			(clear ?from))
	)
\end{lstlisting}
  \end{minipage}
  \caption{Sokoban Domain PDDL (Part 1)}
  \label{lst:sokoban-df-a}
\end{figure*}

\begin{figure*}[t]
  \centering
  \begin{minipage}{\textwidth}
\begin{lstlisting}[language=,basicstyle=\ttfamily\small]
; Sokoban domain (continued)

	(:action push-to-goal
		:parameters (?p - thing ?s - thing ?ppos - location ?from - location ?to - location ?dir - direction)
		:precondition (and (move ?dir)
			(is-player ?p)
			(is-stone ?s)
			(at ?p ?ppos)
			(at ?s ?from)
			(clear ?to)
			(move-dir ?ppos ?from ?dir)
			(move-dir ?from ?to ?dir)
			(is-goal ?to))
		:effect (and
			(not (at ?p ?ppos))
			(not (at ?s ?from))
			(not (clear ?to))
			(at ?p ?from)
			(at ?s ?to)
			(clear ?ppos)
			(at-goal ?s))
	)
	

	(:action push-to-nongoal
		:parameters (?p - thing ?s - thing ?ppos - location ?from - location ?to - location ?dir - direction)
		:precondition (and (move ?dir)
			(is-player ?p)
			(is-stone ?s)
			(at ?p ?ppos)
			(at ?s ?from)
			(clear ?to)
			(move-dir ?ppos ?from ?dir)
			(move-dir ?from ?to ?dir)
			(is-nongoal ?to))
		:effect (and
			(not (at ?p ?ppos))
			(not (at ?s ?from))
			(not (clear ?to))
			(at ?p ?from)
			(at ?s ?to)
			(clear ?ppos)
			(not (at-goal ?s)))
	)

)
\end{lstlisting}
  \end{minipage}
  \caption{Sokoban Domain PDDL (Part 2)}
  \label{lst:sokoban-df-b}
\end{figure*}

\begin{figure*}[t]
  \centering
  \begin{minipage}{\textwidth}
\begin{lstlisting}[language=,basicstyle=\ttfamily\small]
(define (problem sokoban-small)
  (:domain sokoban)
  (:objects
    dir-down - direction
    dir-left - direction
    dir-right - direction
    dir-up - direction
    player-01 - thing
    stone-01 - thing
    pos-1-1 - location
    pos-1-2 - location
    pos-1-3 - location
  )
  (:goal (and (at-goal stone-01)))
  (:init
    (at player-01 pos-1-1)
    (at stone-01 pos-1-2)
    (clear pos-1-3)
    (is-goal pos-1-3)
    (is-nongoal pos-1-1)
    (is-nongoal pos-1-2)
    (move dir-down)
    (move dir-left)
    (move dir-right)
    (move dir-up)
    (is-player player-01)
    (is-stone stone-01)
    (move-dir pos-1-1 pos-1-2 dir-right)
    (move-dir pos-1-2 pos-1-1 dir-left)
    (move-dir pos-1-2 pos-1-3 dir-right)
    (move-dir pos-1-3 pos-1-2 dir-left)
  )
)
\end{lstlisting}
  \end{minipage}
  \caption{Sokoban Problem PDDL}
  \label{lst:sokoban-pf}
\end{figure*}

\begin{figure*}[t]
  \centering
  \begin{minipage}{\textwidth}
\begin{lstlisting}[language=,basicstyle=\ttfamily\small]
1. (push-to-goal player-01 stone-01 pos-1-1 pos-1-2 pos-1-3 dir-right)
\end{lstlisting}
  \end{minipage}
  \caption{Sokoban Plan}
  \label{lst:sokoban-plan}
\end{figure*}

\section{Examples of IRs}
\label{sec:example_ir}

\subsection{NL\textrightarrow PDDL (nl\_pddl)}
Listing \ref{lst:p02-nl_pddl-domain-1} and Listing \ref{lst:p02-nl_pddl-problem-1} present NL-PDDL IR Example.

\begin{figure*}[t]
  \centering
  \begin{minipage}{\textwidth}
\begin{lstlisting}[language=,basicstyle=\ttfamily\small]
Domain Specification BlockStacking
Overview: Arrange blocks into stacks using actions that pick up, unstack, put down, or stack blocks, adhering to constraints on hand state and block clarity.

Entity Types:
- block: Represents individual blocks that can be manipulated.

Predicates:
- HandEmpty: True when no block is being held.
- Holding(b: block): True when block b is in the hand.
- Clear(b: block): True when block b has no blocks on top of it and is not being held.
- OnTable(b: block): True when block b is directly on the table.
- On(b: block, under: block): True when block b is immediately on top of block under.

Actions:
Action: Pickup
Intent: Lift a single block from the table into the hand.
Parameters: b (block)
Preconditions:
- Positive: Clear(b), OnTable(b), HandEmpty
- Negative: None
Effects:
- Add: Holding(b)
- Delete: HandEmpty, OnTable(b), Clear(b)
- Conditional: None

Action: Unstack
Intent: Remove a block from on top of another block and hold it.
Parameters: b (block), under (block)
Preconditions:
- Positive: Clear(b), On(b, under), HandEmpty
- Negative: None
Effects:
- Add: Holding(b), Clear(under)
- Delete: On(b, under), HandEmpty, Clear(b)
- Conditional: None

Action: Putdown
Intent: Place a held block onto the table.
Parameters: b (block)
Preconditions:
- Positive: Holding(b)
- Negative: None
Effects:
- Add: OnTable(b), Clear(b), HandEmpty
- Delete: Holding(b)
- Conditional: None

Action: Stack
Intent: Place a held block onto another block.
Parameters: b (block), under (block)
Preconditions:
- Positive: Holding(b), Clear(under)
- Negative: None
Effects:
- Add: On(b, under), Clear(b), HandEmpty
- Delete: Holding(b), Clear(under)
- Conditional: None
\end{lstlisting}
  \end{minipage}
  \caption{NL-PDDL IR from the NL\textrightarrow PDDL pipeline.}
  \label{lst:p02-nl_pddl-domain-1}
\end{figure*}

\begin{figure*}[t]
  \centering
  \begin{minipage}{\textwidth}
\begin{lstlisting}[language=,basicstyle=\ttfamily\small]
Problem Specification BlockArrangement
Context: Rearrange the blocks from their initial configuration to a specified goal state involving multiple stacks and table placements.

Objects:
- block: 1, 2, 3, 4, 5, 6, 7, 8, 9, 10, 11

Initial State:
- Clear(1) holds
- Clear(2) holds
- Clear(3) holds
- Clear(8) holds
- Clear(9) holds
- Clear(10) holds
- HandEmpty holds
- OnTable(1) holds
- OnTable(2) holds
- OnTable(4) holds
- OnTable(6) holds
- OnTable(8) holds
- OnTable(11) holds
- On(3, 6) holds
- On(5, 4) holds
- On(7, 11) holds
- On(9, 5) holds
- On(10, 7) holds
- Note: Blocks not explicitly listed as On or OnTable (e.g., block 5) are not on the table unless stated. For example, OnTable(5) is false because it is on block 4.

Goal Conditions:
- On(1, 8) must hold
- On(2, 4) must hold
- On(6, 1) must hold
- On(9, 6) must hold
- OnTable(3) must hold
- OnTable(4) must hold
- OnTable(5) must hold
- OnTable(7) must hold
- OnTable(8) must hold
- OnTable(10) must hold
- OnTable(11) must hold
\end{lstlisting}
  \end{minipage}
  \caption{NL-PDDL IR from the NL\textrightarrow PDDL pipeline.}
  \label{lst:p02-nl_pddl-problem-1}
\end{figure*}

\subsection{PythonSim\textrightarrow PDDL (pythonsim\_pddl)}
Listings \ref{lst:p02-pythonsim_pddl-domain-1}--\ref{lst:p02-pythonsim_pddl-domain-3} and Listing \ref{lst:p02-pythonsim_pddl-problem-1} present the PythonSim IR example.

\begin{figure*}[t]
  \centering
  \begin{minipage}{\textwidth}
\begin{lstlisting}[language=Python,basicstyle=\ttfamily\small]
from dataclasses import dataclass, field
from typing import Dict, Set, Optional, Union, Tuple, List, Iterable, frozenset

@dataclass(frozen=True)
class State:
    on: Dict[int, Union[str, int]]  # block -> 'table', another block, or 'held'
    clear: frozenset[int]  # blocks that are clear (no blocks on top and not held)
    held: Optional[int]  # the block being held, or None

class Action:
    def __init__(self, name: str, params: Tuple):
        self.name = name
        self.params = params

    def is_applicable(self, state: State) -> bool:
        raise NotImplementedError

    def apply(self, state: State) -> State:
        raise NotImplementedError

class PickupAction(Action):
    def __init__(self, block: int):
        super().__init__("pickup", (block,))
        self.block = block

    def is_applicable(self, state: State) -> bool:
        b = self.block
        return (state.on[b] == 'table' and
                b in state.clear and
                state.held is None)

    def apply(self, state: State) -> State:
        new_on = state.on.copy()
        new_on[self.block] = 'held'

        new_clear = set(state.clear)
        new_clear.remove(self.block)
        new_clear = frozenset(new_clear)

        new_held = self.block

        return State(new_on, new_clear, new_held)

class UnstackAction(Action):
    def __init__(self, block: int, under: int):
        super().__init__("unstack", (block, under))
        self.block = block
        self.under = under

    def is_applicable(self, state: State) -> bool:
        return (state.on[self.block] == self.under and
                self.block in state.clear and
                state.held is None)

    def apply(self, state: State) -> State:
        new_on = state.on.copy()
        new_on[self.block] = 'held'

        new_clear = set(state.clear)
        new_clear.remove(self.block)
\end{lstlisting}
  \end{minipage}
  \caption{PythonSim IR from the PythonSim\textrightarrow PDDL pipeline (Part 1 of 3).}
  \label{lst:p02-pythonsim_pddl-domain-1}
\end{figure*}

\begin{figure*}[t]
  \centering
  \begin{minipage}{\textwidth}
\begin{lstlisting}[language=Python,basicstyle=\ttfamily\small]
        new_clear.add(self.under)
        new_clear = frozenset(new_clear)

        new_held = self.block

        return State(new_on, new_clear, new_held)

class PutdownAction(Action):
    def __init__(self, block: int):
        super().__init__("putdown", (block,))
        self.block = block

    def is_applicable(self, state: State) -> bool:
        return state.held == self.block

    def apply(self, state: State) -> State:
        new_on = state.on.copy()
        new_on[self.block] = 'table'

        new_clear = set(state.clear)
        new_clear.add(self.block)
        new_clear = frozenset(new_clear)

        new_held = None

        return State(new_on, new_clear, new_held)

class StackAction(Action):
    def __init__(self, block: int, under: int):
        super().__init__("stack", (block, under))
        self.block = block
        self.under = under

    def is_applicable(self, state: State) -> bool:
        return (state.held == self.block and
                self.under in state.clear and
                self.block != self.under)

    def apply(self, state: State) -> State:
        new_on = state.on.copy()
        new_on[self.block] = self.under

        new_clear = set(state.clear)
        new_clear.remove(self.under)
        new_clear.add(self.block)
        new_clear = frozenset(new_clear)

        new_held = None

        return State(new_on, new_clear, new_held)

def available_actions(state: State) -> Iterable[Action]:
    actions = []

    # Pickup
    for b in state.on:
        if state.on[b] == 'table' and b in state.clear and state.held is None:
            actions.append(PickupAction(b))

    # Unstack
\end{lstlisting}
  \end{minipage}
  \caption{PythonSim IR from the PythonSim\textrightarrow PDDL pipeline (Part 2 of 3).}
  \label{lst:p02-pythonsim_pddl-domain-2}
\end{figure*}

\begin{figure*}[t]
  \centering
  \begin{minipage}{\textwidth}
\begin{lstlisting}[language=Python,basicstyle=\ttfamily\small]
    for b in state.on:
        if state.on[b] not in ('table', 'held'):
            under = state.on[b]
            if b in state.clear and state.held is None:
                actions.append(UnstackAction(b, under))

    # Putdown
    if state.held is not None:
        actions.append(PutdownAction(state.held))

    # Stack
    if state.held is not None:
        b = state.held
        for under in state.clear:
            if under != b:
                actions.append(StackAction(b, under))

    return actions

def execute(action: Action, state: State) -> State:
    return action.apply(state)
\end{lstlisting}
  \end{minipage}
  \caption{PythonSim IR from the PythonSim\textrightarrow PDDL pipeline (Part 3 of 3).}
  \label{lst:p02-pythonsim_pddl-domain-3}
\end{figure*}

\begin{figure*}[t]
  \centering
  \begin{minipage}{\textwidth}
\begin{lstlisting}[language=Python,basicstyle=\ttfamily\small]
from typing import List, Tuple

def initial_state() -> State:
    on = {
        1: 'table',
        2: 'table',
        3: 6,
        4: 'table',
        5: 4,
        6: 'table',
        7: 11,
        8: 'table',
        9: 5,
        10: 7,
        11: 'table'
    }
    clear = frozenset({1, 2, 3, 8, 9, 10})
    held = None
    return State(on, clear, held)

def is_goal(state: State) -> bool:
    return (state.held is None and
            state.on[1] == 8 and
            state.on[2] == 4 and
            state.on[6] == 1 and
            state.on[9] == 6 and
            state.on[3] == 'table' and
            state.on[4] == 'table' and
            state.on[5] == 'table' and
            state.on[7] == 'table' and
            state.on[8] == 'table' and
            state.on[10] == 'table' and
            state.on[11] == 'table')

def solve(limit: int = 1024) -> List[Action]:
    from collections import deque

    frontier = deque()
    frontier.append((initial_state(), []))
    visited = set()

    while frontier and len(frontier) < limit:
        state, plan = frontier.popleft()
        if is_goal(state):
            return plan
        if state in visited:
            continue
        visited.add(state)
        for action in available_actions(state):
            next_state = execute(action, state)
            frontier.append((next_state, plan + [action]))
    return []
\end{lstlisting}
  \end{minipage}
  \caption{PythonSim IR from the PythonSim\textrightarrow PDDL pipeline.}
  \label{lst:p02-pythonsim_pddl-problem-1}
\end{figure*}

\subsection{PyPDDL\textrightarrow PDDL (pypddl\_pddl)}
Listings \ref{lst:p02-pypddl_pddl-domain-1}--\ref{lst:p02-pypddl_pddl-domain-2} and Listing \ref{lst:p02-pypddl_pddl-problem-1} present PyPDDL IR example.

\begin{figure*}[t]
  \centering
  \begin{minipage}{\textwidth}
\begin{lstlisting}[language=Python,basicstyle=\ttfamily\small]
from py2pddl import Domain, create_type
from py2pddl import predicate, action

class BlocksWorldDomain(Domain):
    Block = create_type("block")

    @predicate(Block)
    def clear(self, b):
        """Block is clear (no blocks on top and not held)"""

    @predicate(Block, Block)
    def on(self, b, under):
        """Block b is on top of block under"""

    @predicate(Block)
    def on_table(self, b):
        """Block is on the table"""

    @predicate(Block)
    def holding(self, b):
        """Block is being held"""

    @predicate()
    def hand_empty(self):
        """Hand is empty"""

    @action(Block)
    def pickup(self, b):
        precond = [
            self.hand_empty(),
            self.clear(b),
            self.on_table(b)
        ]
        effect = [
            ~self.hand_empty(),
            self.holding(b),
            ~self.on_table(b),
            ~self.clear(b)
        ]
        return precond, effect

    @action(Block, Block)
    def unstack(self, b, under):
        precond = [
            self.hand_empty(),
            self.clear(b),
            self.on(b, under)
        ]
        effect = [
            ~self.hand_empty(),
            self.holding(b),
            ~self.on(b, under),
            ~self.clear(b),
            self.clear(under)
        ]
        return precond, effect

    @action(Block)
    def putdown(self, b):
        precond = [
\end{lstlisting}
  \end{minipage}
  \caption{PyPDDL IR from the PyPDDL\textrightarrow PDDL pipeline (Part 1 of 2).}
  \label{lst:p02-pypddl_pddl-domain-1}
\end{figure*}

\begin{figure*}[t]
  \centering
  \begin{minipage}{\textwidth}
\begin{lstlisting}[language=Python,basicstyle=\ttfamily\small]
            self.holding(b)
        ]
        effect = [
            ~self.holding(b),
            self.hand_empty(),
            self.on_table(b),
            self.clear(b)
        ]
        return precond, effect

    @action(Block, Block)
    def stack(self, b, under):
        precond = [
            self.holding(b),
            self.clear(under)
        ]
        effect = [
            ~self.holding(b),
            self.hand_empty(),
            self.on(b, under),
            ~self.clear(under),
            self.clear(b),
            ~self.on_table(b)
        ]
        return precond, effect
\end{lstlisting}
  \end{minipage}
  \caption{PyPDDL IR from the PyPDDL\textrightarrow PDDL pipeline (Part 2 of 2).}
  \label{lst:p02-pypddl_pddl-domain-2}
\end{figure*}

\begin{figure*}[t]
  \centering
  \begin{minipage}{\textwidth}
\begin{lstlisting}[language=Python,basicstyle=\ttfamily\small]
from py2pddl import init, goal

class BlocksWorldProblem(BlocksWorldDomain):
    def __init__(self):
        super().__init__()
        self.blocks = self.Block.create_objs(range(1, 12), prefix="block")

    @init
    def init(self):
        clear = [
            self.clear(self.blocks[0]),  # block1
            self.clear(self.blocks[1]),  # block2
            self.clear(self.blocks[2]),  # block3
            self.clear(self.blocks[7]),  # block8
            self.clear(self.blocks[8]),  # block9
            self.clear(self.blocks[9]),  # block10
        ]

        hand_empty = [self.hand_empty()]

        on_relations = [
            self.on(self.blocks[2], self.blocks[5]),  # block3 on block6
            self.on(self.blocks[4], self.blocks[3]),  # block5 on block4
            self.on(self.blocks[6], self.blocks[10]), # block7 on block11
            self.on(self.blocks[8], self.blocks[4]),  # block9 on block5
            self.on(self.blocks[9], self.blocks[6]),  # block10 on block7
        ]

        on_table = [
            self.on_table(self.blocks[0]),  # block1
            self.on_table(self.blocks[1]),  # block2
            self.on_table(self.blocks[3]),  # block4
            self.on_table(self.blocks[5]),  # block6
            self.on_table(self.blocks[7]),  # block8
            self.on_table(self.blocks[10]), # block11
        ]

        return clear + hand_empty + on_relations + on_table

    @goal
    def goal(self):
        goals = [
            self.on(self.blocks[0], self.blocks[7]),  # block1 on block8
            self.on(self.blocks[1], self.blocks[3]),  # block2 on block4
            self.on(self.blocks[5], self.blocks[0]),  # block6 on block1
            self.on(self.blocks[8], self.blocks[5]),  # block9 on block6
            self.on_table(self.blocks[2]),            # block3 on table
            self.on_table(self.blocks[3]),            # block4 on table
            self.on_table(self.blocks[4]),            # block5 on table
            self.on_table(self.blocks[6]),            # block7 on table
            self.on_table(self.blocks[7]),            # block8 on table
            self.on_table(self.blocks[9]),            # block10 on table
            self.on_table(self.blocks[10]),           # block11 on table
        ]
        return goals
\end{lstlisting}
  \end{minipage}
  \caption{PyPDDL IR from the PyPDDL\textrightarrow PDDL pipeline.}
  \label{lst:p02-pypddl_pddl-problem-1}
\end{figure*}

\subsection{PDDL\textrightarrow PDDL (pddl\_pddl)}
Listing \ref{lst:p02-pddl_pddl-domain-1} and Listing \ref{lst:p02-pddl_pddl-problem-1} present PDDL IR example.

\begin{figure*}[t]
  \centering
  \begin{minipage}{\textwidth}
\begin{lstlisting}[language=Lisp,basicstyle=\ttfamily\small]
(define
	(domain blocks-world)
	(:requirements :strips :typing)
	(:types block)
	(:predicates
		(clear ?b - block)
		(on ?b - block ?x - block)
		(ontable ?b - block)
		(holding ?b - block)
		(handempty)
	)
	(:action pickup
		:parameters (?b - block)
		:precondition (and (clear ?b) (ontable ?b) (handempty))
		:effect (and (holding ?b) (not (clear ?b)) (not (handempty)) (not (ontable ?b)))
	)
	(:action unstack
		:parameters (?b ?under - block)
		:precondition (and (clear ?b) (on ?b ?under) (handempty))
		:effect (and (holding ?b) (clear ?under) (not (on ?b ?under)) (not (handempty)) (not (clear ?b)))
	)
	(:action putdown
		:parameters (?b - block)
		:precondition (holding ?b)
		:effect (and (ontable ?b) (handempty) (clear ?b) (not (holding ?b)))
	)
	(:action stack
		:parameters (?b ?under - block)
		:precondition (and (holding ?b) (clear ?under))
		:effect (and (on ?b ?under) (handempty) (not (clear ?under)) (not (holding ?b)) (not (ontable ?b)))
	)
)
\end{lstlisting}
  \end{minipage}
  \caption{PDDL IR from the PDDL\textrightarrow PDDL pipeline.}
  \label{lst:p02-pddl_pddl-domain-1}
\end{figure*}

\begin{figure*}[t]
  \centering
  \begin{minipage}{\textwidth}
\begin{lstlisting}[language=Lisp,basicstyle=\ttfamily\small]
(define
	(problem block_problem)
	(:domain blocks-world)
	(:objects block1 block2 block3 block4 block5 block6 block7 block8 block9 block10 block11 - block)
	(:init 
		(clear block1) (clear block2) (clear block3) (clear block8) (clear block9) (clear block10)
		(on block3 block6) (on block5 block4) (on block7 block11) (on block9 block5) (on block10 block7)
		(ontable block1) (ontable block2) (ontable block4) (ontable block6) (ontable block8) (ontable block11)
		(handempty)
	)
	(:goal 
		(and 
			(on block1 block8) (on block2 block4) (on block6 block1) (on block9 block6)
			(ontable block3) (ontable block4) (ontable block5) (ontable block7) (ontable block8) (ontable block10) (ontable block11)
		)
	)
)
\end{lstlisting}
  \end{minipage}
  \caption{PDDL IR from the PDDL\textrightarrow PDDL pipeline.}
  \label{lst:p02-pddl_pddl-problem-1}
\end{figure*}

\section{Examples of Prompts}
\label{sec:example_prompt}

\subsection{Base Prompts}

\begin{figure*}[t]
  \centering
  \begin{minipage}{\textwidth}
\begin{lstlisting}[language={},basicstyle=\ttfamily\scriptsize]
You are an expert automated planner. Your task is to read a natural-language domain description and a corresponding problem description, reason through the objects, actions, and goals, and output a valid plan.

Recall the distinctions:
- The **domain description** explains the general world: available action schemas, predicates, and object types.
- The **problem description** gives the concrete instance: specific objects, the initial state, and the goals for this planning episode.

Planning output requirements:
- Think aloud inside <think>...</think> to lay out objects, preconditions, effects, and progress toward the goals.
- Produce the final plan inside <plan>...</plan>. Each action must be on its own line in classical PDDL plan syntax, e.g. `(move robot1 roomA roomB)`.
- Maintain the execution order from top to bottom. Do not include step numbers, timestamps, or probabilities.
- Only emit actions that are applicable given the initial state and that eventually achieve the goals.
- After the closing </plan> tag, do not add extra commentary.

Example format (replace names with the task-specific details):
<think>
... reasoning about applicable actions ...
</think>
<plan>
(pick-up block1 hand)
(stack block1 block2)
(move robot room1 room2)
</plan>

Follow this template for every problem. EOF
\end{lstlisting}
  \end{minipage}
  \caption{Base instruction \texttt{plan\_instruction}.}
  \label{lst:plan_instruction}
\end{figure*}

\begin{figure*}[t]
  \centering
  \begin{minipage}{\textwidth}
\begin{lstlisting}[language={},basicstyle=\ttfamily\scriptsize]
PDDL domain file contains domain name, requirements, types of objects in the domain, predicates, and actions.
Based on the natural language domain description, identify the actions that are possible. 
Identify action sematics i.e. understand the preconditions under which that action could be done and the effects of the action.
Then identify appropriate predicates that could enable action semantics i.e. preconditions and effects.
PDDL domain file has a definitive syntax that must be followed for any domain. An abstract example PDDL domain file is given below:

<domain_file>
(define
	(domain domain_name)
	(:requirements :strips :typing)
	(:types
		type1
		type2
	)
	(:predicates
		(predicate1 ?arg1 - type1 ?arg2 - type2)
		(predicate2 ?arg1 - type1 ?arg2 - type2)
	)
	(:action action1
		:parameters (?arg1 - type1 ?arg2 - type2 ?arg3 - type2)
		:precondition (predicate1 ?arg1 ?arg2)
		:effect (and (predicate1 ?arg1 ?arg2) (predicate2 ?arg1 ?arg3))
	)
	(:action action2
		:parameters (?arg1 - type1 ?arg2 - type2 ?arg3 - type2)
		:precondition (and (predicate1 ?arg1 ?arg2) (predicate2 ?arg1 ?arg3))
		:effect (predicate2 ?arg1 ?arg3)
	)
)
</domain_file>

Notes for generating domain file: 
- type1 & type2 are only representative and should be replaced with appropriate types. There could be any number of types.
- predicate1 & predicate2 are only representative and should be replaced with appropriate predicates. There could be any number of predicates.
- action1 & action2 are only representative and should be replaced with appropriate actions. There could be any number of actions.
- arg1 & arg2 are only representative and should be replaced with appropriate arguments for predicates and in preconditions and effects.
- predicates with proper arguments could be combined to combine complex boolean expression to represent predicondition and effect 
The braces should be balanced for each section of the PDDL program
- Use predicates with arguments of the right type as declared in domain file
- All the arguments to any :precondition or :effect of an action should be declared in :parameters as input arguments


PDDL problem file contains problem name, domain name, objects in this problem instance, init state of objects, and goal state of objects.
Based on the natural language problem description, identify the relevant objects for this problems with their names and types.
Represent the initial state with the appropriate predicates and object arguments. Represent the goal state with the appropriate predicates and object arguments.
PDDL problem file has a definitive syntax that must be followed for any problem. An abstract example PDDL problem file is given below.

<problem_file>
(define
	(problem problem_name)
	(:domain domain_name)
	(:objects
		obj1 obj2 - type1
		obj3, obj4 - type2
	)
	(:init (predicate1 obj1 obj3) (predicate2 obj2 obj3))
	(:goal (and (predicate1 obj1 obj4) (predicate2 obj2 obj3)))
)
</problem_file>

Notes for generating problem file:
- obj1, obj2, ... are only representative and should be replaced with appropriate objects. There could be any number of obects with their types.
- init state with predicate1 & predicate2 is only representative and should be replaced with appropriate predicates that define init state
- goal state with predicate1 & predicate2 is only representative and should be replaced with appropriate predicates that define goal state
- predicates with proper arguments could be combined to combine complex boolean expression to represent init and goal states 
- The braces should be balanced for each section of the PDDL program
- Use predicates with arguments of the right type as declared in domain file
- All the objects that would be arguments of predicates in init and goal states should be declared in :objects
\end{lstlisting}
  \end{minipage}
  \caption{Base instruction \texttt{only\_pddl\_instruction}.}
  \label{lst:only_pddl_instruction}
\end{figure*}

\begin{figure*}[t]
  \centering
  \begin{minipage}{\textwidth}
\begin{lstlisting}[language={},basicstyle=\ttfamily\scriptsize]
Natural language planning specifications serve as a blueprint for generating the exact PDDL domain and problem files. Write in clear prose, but ensure every detail required for a faithful PDDL reconstruction is present and testable.

Overall guidance:
- Organize the specification with labeled sections so the structure is obvious without PDDL syntax.
- Use precise terminology and consistent naming; anything omitted here cannot appear in the final PDDL.
- State facts, conditions, and effects explicitly. Avoid implication, ambiguity, or narrative fluff.

Domain description must include:
- Domain title and a one-sentence overview of the scenario.
- Complete list of entity types, subtyping relations when relevant, and the role each type plays.
- Exhaustive catalog of predicates. For each predicate, name every parameter, state its type, and clarify exactly what the predicate means when true.
- Exhaustive catalog of actions. For each action, provide:
  * Action title and concise intent.
  * Parameter list with names and types, including any constraints or distinctness requirements.
  * Preconditions broken out as positive conditions (facts that must hold) and negative conditions (facts that must not hold), both referencing the predicates above.
  * Effects split into add effects (facts that become true) and delete effects (facts that cease to hold). Mention conditional effects separately if they exist.
  * Any invariants or notes needed to enforce the same logical behavior as PDDL.

Problem description must include:
- Problem title and one-sentence context linking it to the domain.
- Exhaustive list of objects, grouped by type, using the exact type names from the domain section.
- Initial state described as bullet points or sentences, each mapping directly to a predicate instance that should hold initially. Include explicit mention of facts that are false if the domain relies on them.
- Goal conditions stated explicitly, each tied to predicate instances. Distinguish between conjunctive and disjunctive goals if applicable.

Formatting conventions:
- Stick to full sentences or crisp bullet points-no raw PDDL, no code blocks, no decorative markup.
- Maintain consistent vocabulary between domain and problem sections so predicates, actions, and objects align exactly.
- If numeric fluents, temporal elements, or other advanced PDDL features are present, describe their roles, bounds, and update rules explicitly.
- The specification must be sufficient for a separate model to regenerate the original PDDL without guessing.

Abstract example (replace placeholders with the concrete content for the task at hand):
Domain Specification - DomainName
Overview: A single sentence summarizing the planning scenario.
Entity Types:
- Type1: Explanation of entities of this type and any subtype relationships.
- Type2: Explanation of entities of this type.
Predicates:
- predicate1(entity1: Type1, entity2: Type2): Meaning of the fact when true.
- predicate2(entity: Type2): Meaning of the fact when true.
Actions:
Action: Action1
Intent: Brief purpose of the action.
Parameters: entity1: Type1, entity2: Type2 (include distinctness constraints if required).
Preconditions:
- Positive: predicate1(entity1, entity2) must hold.
- Negative: predicate2(entity2) must not hold.
Effects:
- Add: predicate2(entity2) becomes true.
- Delete: predicate1(entity1, entity2) ceases to hold.
- Conditional: Describe any conditional effects explicitly, including their triggering conditions.
Action: Action2
Intent: Another representative action with its parameters, preconditions, and effects fully spelled out.

Problem Specification - ProblemName
Context: One sentence linking the problem to the domain.
Objects:
- Type1: obj1, obj2 (brief descriptions optional).
- Type2: obj3, obj4.
Initial State:
- predicate1(obj1, obj3) holds because ...
- predicate2(obj4) is false (state explicitly if the domain depends on this).
Goal Conditions:
- Conjunctive goal requiring predicate2(obj3) and predicate2(obj4) to hold.
- Note disjunctions or temporal ordering if the problem includes them.
\end{lstlisting}
  \end{minipage}
  \caption{Base instruction \texttt{nl\_instruction}.}
  \label{lst:nl_instruction}
\end{figure*}

\subsection{Plan Generation (plan\_gen)}

Uses \texttt{plan\_instruction} together with the natural-language domain and problem descriptions for direct plan synthesis.

\begin{figure*}[t]
  \centering
  \begin{minipage}{\textwidth}
\begin{lstlisting}[language={},basicstyle=\ttfamily\scriptsize]
{{plan_instruction}}

### Domain description
{{domain_description}}

### Problem description
{{problem_description}}
Generate the plan for this instance following the required format.
\end{lstlisting}
  \end{minipage}
  \caption{Prompt template for \texttt{plan\_gen} plan generation.}
  \label{lst:plan_gen_prompt}
\end{figure*}

\subsection{Single-Stage PDDL (pddl)}

Uses \texttt{only\_pddl\_instruction} before asking for domain and problem files grounded in the original descriptions.

\begin{figure*}[t]
  \centering
  \begin{minipage}{\textwidth}
\begin{lstlisting}[language={},basicstyle=\ttfamily\scriptsize]
{{only_pddl_instruction}}

You are an expert PDDL engineer.
Create syntactically correct and mutually consistent domain and problem files using the information below.

### Original domain description
{{domain_description}}

### Original problem description
{{problem_description}}

Wrap only the final domain PDDL in <domain_file>...</domain_file> and the final problem PDDL in <problem_file>...</problem_file>.
\end{lstlisting}
  \end{minipage}
  \caption{Prompt template for \texttt{pddl}.}
  \label{lst:pddl_prompt}
\end{figure*}

\subsection{NL\texorpdfstring{$\rightarrow$}{→}PDDL (nl\_pddl)}

Stage one converts natural language to a structured prose specification using \texttt{nl\_instruction}; stage two calls \texttt{only\_pddl\_instruction} to emit PDDL from the summary plus original descriptions.

\begin{figure*}[t]
  \centering
  \begin{minipage}{\textwidth}
\begin{lstlisting}[language={},basicstyle=\ttfamily\scriptsize]
{{nl_instruction}}

Use the following natural-language descriptions to craft a complete planning specification in prose. Follow the layout described above.

### Original domain description
{{domain_description}}

### Original problem description
{{problem_description}}

Wrap the final structured prose in <nl_summary>...</nl_summary>. Begin your reasoning with <think> so we can inspect the chain of thought.
Do not emit any PDDL or code yet.
\end{lstlisting}
  \end{minipage}
  \caption{Stage-one prompt for \texttt{nl\_pddl}.}
  \label{lst:nl_pddl_stage1}
\end{figure*}

\begin{figure*}[t]
  \centering
  \begin{minipage}{\textwidth}
\begin{lstlisting}[language={},basicstyle=\ttfamily\scriptsize]
{{only_pddl_instruction}}

You are an expert PDDL engineer.
Leverage the structured natural-language planning specification, along with the original descriptions, to produce consistent domain and problem files. Always ground your final answer in the original domain and problem descriptions shown below.

### Original domain description
{{domain_description}}

### Original problem description
{{problem_description}}

### Natural-language planning specification
{{nl_summary}}

Output syntactically correct PDDL. Wrap only the domain file in <domain_file>...</domain_file> and the problem file in <problem_file>...</problem_file>.
\end{lstlisting}
  \end{minipage}
  \caption{Stage-two prompt for \texttt{nl\_pddl}.}
  \label{lst:nl_pddl_stage2}
\end{figure*}

\subsection{PythonSim\texorpdfstring{$\rightarrow$}{→}PDDL (pythonsim\_pddl)}

Stage one gathers a Python simulator via \texttt{python\_sim\_instruction}; stage two invokes \texttt{only\_pddl\_instruction} with the simulator alongside the original descriptions.

\begin{figure*}[t]
  \centering
  \begin{minipage}{\textwidth}
\begin{lstlisting}[language={},basicstyle=\ttfamily\scriptsize]
{{python_sim_instruction}}

### Domain description
{{domain_description}}

### Problem description
{{problem_description}}

Synthesise the Python planning simulator. Wrap the reusable domain code in <domain_file>...</domain_file> and the problem-specific code in <problem_file>...</problem_file>.
\end{lstlisting}
  \end{minipage}
  \caption{Stage-one prompt for \texttt{pythonsim\_pddl}.}
  \label{lst:pythonsim_pddl_stage1}
\end{figure*}

\begin{figure*}[t]
  \centering
  \begin{minipage}{\textwidth}
\begin{lstlisting}[language={},basicstyle=\ttfamily\scriptsize]
{{only_pddl_instruction}}

You are an expert PDDL engineer. Using the natural-language descriptions and the Python simulator below, produce syntactically correct and mutually consistent domain/problem PDDL files. Your answer must stay aligned with the original domain and problem descriptions referenced above.

### Original domain description
{{domain_description}}

### Original problem description
{{problem_description}}

### Python simulator \- domain module
```python
{{python_domain_module}}
```

### Python simulator \- problem module
```python
{{python_problem_module}}
```

Wrap only the final domain PDDL in <domain_file>...</domain_file> and the final problem PDDL in <problem_file>...</problem_file>.
\end{lstlisting}
  \end{minipage}
  \caption{Stage-two prompt for \texttt{pythonsim\_pddl}.}
  \label{lst:pythonsim_pddl_stage2}
\end{figure*}

\subsection{PyPDDL\texorpdfstring{$\rightarrow$}{→}PDDL (pypddl\_pddl)}

Stage one elicits PyPDDL code with \texttt{pypddl\_instruction}; stage two relies on \texttt{only\_pddl\_instruction} plus the generated program.

\begin{figure*}[t]
  \centering
  \begin{minipage}{\textwidth}
\begin{lstlisting}[language={},basicstyle=\ttfamily\scriptsize]
{{pypddl_instruction}}

### Domain description
{{domain_description}}

### Problem description
{{problem_description}}

Write the PyPDDL domain and problem classes. Wrap the domain in <domain_file>...</domain_file> and the problem in <problem_file>...</problem_file>.
\end{lstlisting}
  \end{minipage}
  \caption{Stage-one prompt for \texttt{pypddl\_pddl}.}
  \label{lst:pypddl_pddl_stage1}
\end{figure*}

\begin{figure*}[t]
  \centering
  \begin{minipage}{\textwidth}
\begin{lstlisting}[language={},basicstyle=\ttfamily\scriptsize]
{{only_pddl_instruction}}

You are an expert PDDL engineer. Using the natural-language descriptions and the PyPDDL program below, produce syntactically correct, mutually consistent domain and problem PDDL files. Keep the original descriptions in mind throughout your answer.

### Original domain description
{{domain_description}}

### Original problem description
{{problem_description}}

### PyPDDL domain class
```python
{{pypddl_domain_code}}
```

### PyPDDL problem class
```python
{{pypddl_problem_code}}
```

Wrap only the final domain PDDL in <domain_file>...</domain_file> and the final problem PDDL in <problem_file>...</problem_file>.
\end{lstlisting}
  \end{minipage}
  \caption{Stage-two prompt for \texttt{pypddl\_pddl}.}
  \label{lst:pypddl_pddl_stage2}
\end{figure*}

\subsection{PDDL\texorpdfstring{$\rightarrow$}{→}PDDL with Feedback (pddl\_pddl)}

Stage one uses \texttt{pddl\_instruction} for direct PDDL; stage two revises the files using solver feedback from the first attempt.

\begin{figure*}[t]
  \centering
  \begin{minipage}{\textwidth}
\begin{lstlisting}[language={},basicstyle=\ttfamily\scriptsize]
{{pddl_instruction}}

Domain description:
{{domain_description}}

Problem description:
{{problem_description}}
Write the domain and problem files in minimal PDDL.

Wrap the domain inside <domain_file>...</domain_file> and the problem inside <problem_file>...</problem_file>.
\end{lstlisting}
  \end{minipage}
  \caption{Stage-one prompt for \texttt{pddl\_pddl}.}
  \label{lst:pddl_pddl_stage1}
\end{figure*}

\begin{figure*}[t]
  \centering
  \begin{minipage}{\textwidth}
\begin{lstlisting}[language={},basicstyle=\ttfamily\scriptsize]
{{pddl_instruction}}

Domain description:
{{domain_description}}

Problem description:
{{problem_description}}
Write the domain and problem files in minimal PDDL.

To aid the revision, the original natural-language descriptions are repeated above.
------ Previous attempt start ------
[DOMAIN FILE]
{{stage1_domain_file}}

[PROBLEM FILE]
{{stage1_problem_file}}
------ Previous attempt end --------

Feedback from the planning solver:
{{stage1_solver_feedback}}

Revise the domain and problem to fix the issues. Output only the new domain wrapped in <domain_file>...</domain_file> and the new problem wrapped in <problem_file>...</problem_file>.
\end{lstlisting}
  \end{minipage}
  \caption{Stage-two prompt for \texttt{pddl\_pddl}.}
  \label{lst:pddl_pddl_stage2}
\end{figure*}

\subsection{PDDL\texorpdfstring{$\rightarrow$}{→}PDDL\texorpdfstring{$\rightarrow$}{→}PDDL (pddl\_pddl\_pddl)}

Applies \texttt{pddl\_instruction} across three solver-informed revisions.

\begin{figure*}[t]
  \centering
  \begin{minipage}{\textwidth}
\begin{lstlisting}[language={},basicstyle=\ttfamily\scriptsize]
{{pddl_instruction}}

Domain description:
{{domain_description}}

Problem description:
{{problem_description}}
Write the domain and problem files in minimal PDDL.

Wrap the domain inside <domain_file>...</domain_file> and the problem inside <problem_file>...</problem_file>.
\end{lstlisting}
  \end{minipage}
  \caption{Stage-one prompt for \texttt{pddl\_pddl\_pddl}.}
  \label{lst:pddl_pddl_pddl_stage1}
\end{figure*}

\begin{figure*}[t]
  \centering
  \begin{minipage}{\textwidth}
\begin{lstlisting}[language={},basicstyle=\ttfamily\scriptsize]
{{pddl_instruction}}

Domain description:
{{domain_description}}

Problem description:
{{problem_description}}
Write the domain and problem files in minimal PDDL.

To aid the revision, the original natural-language descriptions are repeated above.
------ Previous attempt start ------
[DOMAIN FILE]
{{stage1_domain_file}}

[PROBLEM FILE]
{{stage1_problem_file}}
------ Previous attempt end --------

Solver feedback:
{{stage1_solver_feedback}}

Revise the domain and problem to fix the issues. Output only the new domain wrapped in <domain_file>...</domain_file> and the new problem wrapped in <problem_file>...</problem_file>.
\end{lstlisting}
  \end{minipage}
  \caption{Stage-two prompt for \texttt{pddl\_pddl\_pddl}.}
  \label{lst:pddl_pddl_pddl_stage2}
\end{figure*}

\begin{figure*}[t]
  \centering
  \begin{minipage}{\textwidth}
\begin{lstlisting}[language={},basicstyle=\ttfamily\scriptsize]
{{pddl_instruction}}

Domain description:
{{domain_description}}

Problem description:
{{problem_description}}
Write the domain and problem files in minimal PDDL.

Review the latest PDDL below together with the solver feedback and produce a final corrected version.

### Previous domain.pddl
```pddl
{{stage2_domain_file}}
```

### Previous problem.pddl
```pddl
{{stage2_problem_file}}
```

### Solver feedback
{{stage2_solver_feedback}}

Wrap the refined domain in <domain_file>...</domain_file> and the refined problem in <problem_file>...</problem_file>.
\end{lstlisting}
  \end{minipage}
  \caption{Stage-three prompt for \texttt{pddl\_pddl\_pddl}.}
  \label{lst:pddl_pddl_pddl_stage3}
\end{figure*}

\end{document}